\documentclass{article}

\usepackage{microtype}
\usepackage{graphicx}
\usepackage{subfigure}
\usepackage{booktabs} 

\usepackage{hyperref}

\usepackage[accepted]{icml2021}

\usepackage{multirow}
\usepackage{dblfloatfix} 
\usepackage{bm}

\newtheorem{theorem}{Theorem}

\usepackage{amsmath,amsfonts}
\usepackage{amssymb} 
\usepackage{dsfont} 
\usepackage{enumitem}

\newcommand{\cC}{\mathcal{C}}
\newcommand{\cD}{\mathcal{D}}

\newcommand{\cO}{\mathcal{O}}
\newcommand{\cT}{\mathcal{T}}
\newcommand{\cV}{\mathcal{V}}
\allowdisplaybreaks

\newcommand{\myblue}[1]{#1}
\usepackage{commands}

\icmltitlerunning{Optimal Counterfactual Explanations in Tree Ensembles}

\begin{document}

\twocolumn[
\icmltitle{Optimal Counterfactual Explanations in Tree Ensembles}



\icmlsetsymbol{equal}{*}

\begin{icmlauthorlist}
\icmlauthor{Axel Parmentier}{ponts}
\icmlauthor{Thibaut Vidal}{poly,puc}
\end{icmlauthorlist}

\icmlaffiliation{ponts}{\textsc{Cermics}, \'Ecole des Ponts Paristech;}
\icmlaffiliation{puc}{Department of Computer Science, Pontifical Catholic University of Rio de Janeiro (PUC-Rio), Brazil}
\icmlaffiliation{poly}{CIRRELT \& SCALE-AI Chair in Data-Driven Supply Chains, Department of Mathematics and Industrial Engineering, Polytechnique Montreal, Canada;}

\icmlcorrespondingauthor{Thibaut Vidal}{thibaut.vidal@cirrelt.ca}

\icmlkeywords{Accountability, Interpretability, Counterfactual Explainations, Tree Ensembles}

\vskip 0.3in
]



\printAffiliationsAndNotice{}  

\begin{abstract}
Counterfactual explanations are usually generated through heuristics that are sensitive to the search's initial conditions. The absence of guarantees of performance and robustness hinders trustworthiness. In this paper, we take a disciplined approach towards counterfactual explanations for tree ensembles. We advocate for a model-based search aiming at ``optimal'' explanations and propose efficient mixed-integer programming approaches. We show that isolation forests can be modeled within our framework to focus the search on plausible explanations with a low outlier score. We provide comprehensive coverage of additional constraints that model important objectives, heterogeneous data types, structural constraints on the feature space, along with resource and actionability restrictions. Our experimental analyses demonstrate that the proposed search approach requires a computational effort that is orders of magnitude smaller than previous mathematical programming algorithms. It scales up to large data sets and tree ensembles, where it provides, within seconds, systematic explanations grounded on well-defined models solved to optimality.
\end{abstract}

\section{Introduction}
\label{sec:Introduction}

Accountability in machine learning is quickly rising as a major concern as learning algorithms take over tasks that have a major impact on human lives. With the increasing use of profiling and automated decision-making systems, new legal provisions are being set up to protect rights to transparency. The recent interpretation of the EU General Data Protection Regulation (GDPR) by \citet{WorkingParty2018} refers to a ``right to explanations'' and has triggered extensive research on algorithmic recourse and counterfactual explanations \citep[see, e.g.,][]{Wachter2018,Karimi2020,Verma2020}.

Counterfactual explanations are contrastive arguments of the type: ``To obtain this loan, you need \$40,000 of annual revenue instead of the current \$30,000''. They correspond to small perturbations of an example that permit to modify the classification outcome, in a similar fashion as adversarial examples, though typically restricted by additional constraints ensuring actionability and plausibility \citep{Barocas2020,Venkatasubramanian2020}.

Despite their conceptual simplicity, counterfactual explanations pose significant challenges related to data protection, intellectual property, ethics, along with fundamental computational tractability issues.
Indeed, the scale of machine learning models has tremendously increased over a few decades. Even when restricted to the vicinity of an example, a systematic inspection of all possible explanations is intractable, and therefore most studies on counterfactual explanations rely on ad-hoc algorithms or heuristics (e.g., gradient descent in a non-convex space in \citealt{Wachter2018}). This poses at least three main issues:
\begin{enumerate}
    \item[(a)] Heuristics can fail to identify the most natural and insightful explanation, and therefore do not necessarily give a trustworthy cause \citep{Karimi2020a}. This is especially true when the search involves combinatorial spaces (e.g., tree ensembles) with binary or integer features bound together by plausibility constraints.
    \item[(b)] They can be sensitive to the initial conditions of the search, leading to unstable results ---even for the same subject.
    \item[(c)]  Finally, these methods are not readily extensible to include additional constraints and domain knowledge regarding actionability and plausibility. A small change of problem formulation due to a specific application domain can very well require significant methodological adaptations.
\end{enumerate}

To circumvent these issues, we advocate for a disciplined analysis of counterfactual explanations through mathematical programming lenses. We focus on tree ensembles (including random forests and gradient boosting), a popular family of models with good empirical performance which is often sought as a more transparent replacement to neural networks \citep{Rudin2019}. 
We opt for a solution approach grounded on mixed integer linear and quadratic programming (MILP and MIQP) since we believe that it solves the three aforementioned issues. Firstly, the search for an optimal solution 
of a well-defined model permits to control the quality of the solution and ensures the stability of the results, solving key issues (a) and~(b). Moreover, as seen later in this paper, the modeling capacities of MILP permit to seamlessly integrate domain information as well as many forms of plausibility and actionability constraints, therefore making meaningful progress on issue (c). Finally, the tremendous progress of MILP solution approaches over the years (estimated to a $10^{11}$ reduction of computational effort on the same problems -- \citealt{Bixby2012}) now permits to use such a modeling and search approach to its full potential. 

\textbf{We make the following contributions:}\vspace*{-0.25cm}
\begin{enumerate}
    \item We propose the first efficient mathematical models to search for counterfactual explanations in tree ensembles with a number of binary variables that is logarithmic in the number of vertices, and therefore scales to large trees while retaining its ability to find an optimal solution. This is exponentially fewer variables than the previous models used in \citet{Cui2015} and \citet{Kanamori2020}. Our approach is applicable to heterogeneous datasets with numerical, ordinal, binary, and categorical features, with possible oblique splits on numerical features, and considering single or multiple classes. In contrast with previous works, it does not require binary variables to model the numerical feature levels and has a sparse constraint matrix. Consequently, solution performance remains stable as the number of features increases.
    \item We demonstrate how to integrate plausibility in our mathematical framework from an isolation forest viewpoint. Isolation forests are an effective and distribution-agnostic way to associate a plausibility score for the different regions of the feature space. This is, to our knowledge, the first time that this framework is used for counterfactual explanations, through an integration which is only possible due to our significant model-tractability improvements.
    \item We discuss extensions of the model capturing important constraints regarding plausibility and actionability. 
    We therefore provide a flexible and modular toolset that can be adapted to each specific situation.
    \item Finally, we conduct an extensive and reproducible experimental campaign, which can be executed from a single self-contained Python script. Our source code is openly accessible at \url{https://github.com/vidalt/OCEAN} under a MIT license. We demonstrate that the approaches proposed in this work are efficient and scalable, producing \emph{optimal} counterfactual explanations in a matter of a few seconds on data sets with over fifty features, using tree ensembles with hundreds of trees. We evaluate the impact of our plausibility constraints via isolation forests, demonstrating the flexibility of the approach and showing that these extra constraints do not significantly impact the performance of the solution process while significantly boosting the usefulness of the explanations.
\end{enumerate}


\section{Background}
\label{sec:Related-Works}

\subsection{Mixed Integer Programming}

MIQPs can be cast into the following standard form:
\begin{align}
\min \hspace*{0.2cm} & f(\mathbf{x}) \label{MIPgeneral:obj} \\
\text{s.t. } & A \mathbf{x} \leq  \mathbf{b} \label{MIPgeneral:poly} \\
&  \mathbf{x}^\intercal Q_i \mathbf{x} + \mathbf{c}_i \mathbf{x} \leq b_i & i \in \{1,\dots, m\} \label{MIPgeneral:quad} \\
& \mathbf{x} \in \smash{ \mathbb{Z}^a \times \mathbb{R}^b},
\end{align}
where $a$ represents the number of variables taking integer values and $b$ is the number of continuous variables. The feasibility region of the problem is defined as the intersection of a polytope (Constraint \ref{MIPgeneral:poly}) along with a set of quadratic restrictions (Constraint \ref{MIPgeneral:quad}).
State-of-the-art solvers (e.g., CPLEX and Gurobi) can handle separable quadratic objectives of the form $f(\mathbf{x}) = \mathbf{x}^\intercal Q \mathbf{x} + \mathbf{c} \mathbf{x}$ and therefore model a wide range of objectives (regularization terms through $l_0$, $l_1$ and squared $l_2$ norms, squared Euclidean and Mahalanobis distances, and variations thereof). MILP and MIQP are NP-hard in general, though astonishing progress in solution methods has permitted to handle increasingly large problems. Solver performance is, however, dependent on the quality of the problem formulation (i.e., the model). Ideally, a good model should have few binary variables, limited symmetry, and a strong continuous relaxation, i.e., a small gap between its optimal solution value and that of the same problem in which variables $\mathbf{x}$ are relaxed to the domain~$\mathbb{R}^{a+b}$, as this permits to quickly prune regions of the search space during the branch-and-bound process \citep{Wolsey2020}.

\subsection{Counterfactual Explanations in Tree Ensembles}

Let $\{\mathbf{x}_k,c_k\}_{k=1}^n$ be a training set in which each \mbox{$\mathbf{x}_k \in \mathbb{R}^p$} corresponds to a sample characterized by a $p$-dimensional feature vector and a class $c_k \in \cC$. In the most general form, a tree ensemble~$\mathcal{T}$ learns a set of trees $t \in \mathcal{T}$ returning class probabilities $\smash{F_{tc}:\mathcal{X} \rightarrow [0,1]}$. For any sample~$\mathbf{x}$, the tree ensemble returns the class $c$ that maximizes the weighted sum of the probabilities: $F_\mathcal{T}(\mathbf{x}) = \argmax_c \sum w_t F_{tc} (\mathbf{x})$. 
Given an origin point $\hat{\mathbf{x}}$ and a desired prediction class~$c^*$, searching for a plausible and actionable counterfactual explanation consists in locating a new data point $\mathbf{x} \in \mathcal{X}$ that solves the following problem:
\begin{align}
\min \hspace*{0.2cm} & f_{\hat{\mathbf{x}}}(\mathbf{x}) \label{CE:cost} \\
\text{s.t. } \hspace*{0.2cm} & F_\mathcal{T}(\mathbf{x}) = c^* \label{CE:forest} \\
& \smash{\mathbf{x} \in X^\textsc{p} \cap X^\textsc{a}}. \label{CE:domain}
\end{align}
In this problem, $f_{\hat{\mathbf{x}}}$ is a separable convex cost 
that represents how difficult it is to move from $\hat{\mathbf{x}}$ to $\mathbf{x}$. This generic cost function includes distance metrics (e.g., squared Euclidean or Mahalanobis distance) as a special case. Moreover, it allows possible cost asymmetry \citep{Ustun2019,Karimi2020} and can include additional penalization terms if needed. Polytopes $X^\textsc{p}$ and $X^\textsc{a}$ represent the space of plausible and actionable counterfactual explanations, and will be discussed in the next paragraphs in connection with recent works.

\textbf{Related studies}
As reviewed in \citet{Guidotti2018}, \citet{Karimi2020} and \citet{Verma2020}, early studies on counterfactual explanations were conducted in majority in a model-agnostic context through enumeration and heuristic search approaches \citep{Wachter2018}. Dedicated work has been later conducted on specific models, such as tree ensembles, which presented additional challenges due to their combinatorial and non-differentiable nature. To handle this case, \citet{Lucic2019} proposed a gradient algorithm that approximates the splits of the decision trees through sigmoid functions. \citet{Tolomei2017a} designed a feature tweaking algorithm that enumerates alternative paths in each tree to change the decision of the ensemble. The method is shown to provide useful counterfactual explanations on several application cases, though it does not always deliver an optimal (or even a feasible) counterfactual explanation for tree ensembles. To circumvent these issues, \citet{Karimi2020a} reformulated the search for counterfactual explanations in heterogeneous domains (with binary, numerical and categorical features) as a satisfiability problem and employed specialized solvers. The authors reported promising results on three data sets involving up to $14$ features.

\citet{Cui2015} and \citet{Kanamori2020} proposed mixed-integer linear programming approaches for optimal explanations in tree ensembles. The former work considers Mahalanobis distance and introduced decision logic constraints to ensure the consistency of the split decisions through the forest (ensuring that there is exists a counterfactual example satisfying them). The later work demonstrated how to expand the formulation to consider an $l_1$-norm distance and plausibility constraints grounded on the Local Outlier Factor (LOF) score. Both models use a discretization of the feature space, and therefore need a large number of binary variables to express continuous features (one for each possible feature level, and one for each leaf of each decision tree). Good results were still achieved on a variety of data sets. As demonstrated in our work, better formulations relying on exponentially fewer variables can be designed, leading to reductions of CPU time by some orders of magnitude. Finally, a few other works related to mathematical programming do not necessarily consider tree ensembles explanations but provide useful additional modeling strategies. In particular, \citet{Russell2019} modeled the choice of a feature level among multiple intervals (a special case of disjunctive constraint discussed in \citealt{Jeroslow1984}) and suggest strategies to find multiple explanations. Moreover, \citet{Ustun2019} propose to improve actionability through the definition of extra constraints representing a partial causal model.

\textbf{Plausibility and actionability.}
Constraint~(\ref{CE:domain}) aims at ensuring plausibility and actionability of the counterfactual explanations. These important requirements constitute one of the cruces of recent research on counterfactual explanations \citep{Barocas2020}. Plausibility constraints (polytope $X^\textsc{p}$) should ensure that the explanation $\mathbf{x}$ respects the structure of the data and that it is located in a region that has a sufficiently large density of samples. To capture this notion, we will rely on the information of isolation forests \citep{Liu2008a} within our framework to restrict the search to plausible regions. In contrast, actionability constraints (polytope $X^\textsc{a}$) concern the trajectory between $\hat{\mathbf{x}}$ and $\mathbf{x}$. At the very least, they ensure that immutable features remain fixed and that features that are bound to evolve unilaterally are constrained to remain within a half-space. A finer-grained knowledge of correlations (or even a partial causal model as discussed in \citealt{Mahajan2019}, \citealt{Ustun2019}, \citealt{Mothilal2020} and \citealt{Karimi2020b}) can also be integrated into the formulation through additional linear and logical constraints, as discussed in Section~\ref{sec:model4}.

\section{Methodology}
\label{sec:Methodology}

Our mathematical model is presented in two main stages. First, we describe the variables and constraints that characterize the branches taken by the counterfactual example. Next, we include additional variables and constraints modeling the counterfactual example's feature values and ensuring compatibility with all the branch choices.

Our formulation relies on two main pillars. First, it uses the natural disjunctive structure of the trees to model branch choices using exponentially fewer binary variables than previous models by \citet{Cui2015} and \citet{Kanamori2020}. Second, it includes continuous variables organized as order simplices \citep{Grotzinger1984} to represent the values of numerical or ordinal features of the counterfactual example and to connect them with the branch choices. This effectively leads to a formulation requiring only $\cO(N_v)$ non-zero terms in the constraint matrix instead of $\cO(N_v^2)$, where~$N_v$ stands for the overall number of internal nodes in the tree ensemble. Notably, this formulation complexity does not depend on the number of features of the data set. Our model is especially suitable for large data sets with many numerical features, and permits us to rely on isolation forests to model plausibility without sacrificing numerical tractability.

\subsection{Branch Choices}
\label{sec:model1}

For each tree $t \in \cT$, let $\cV^I_t$ be the set of internal vertices associated with the splits, and let $\cV^L_t$ be the set of terminal vertices (leaves). Let $\cD_t$ represent the possible depths values and define $\cV^I_{td}$ as the set of internal nodes at depth $d$.
Let $l(v)$ and $r(v)$ be the left and right children of each internal vertex $v \in \cV^I_t$. Finally, let $p_{tvc}$ be the class probability of $c \in \cC$ in each leaf $v \in \cV^L_t$. Class probabilities can be defined in~$\{0,1\}$ in the hard voting model (as in the random forest algorithm initially proposed by \citealt{Breiman2001}), or in $[0,1]$ in the general soft voting model (e.g., as in scikit-learn).

For each tree $t$ and depth $d$ in $\cD_t$, we use a binary decision variable $\lambda_{td}$ which will take value $1$ if the counterfactual example descends towards the left branch, and $0$ otherwise. This value is free if the path does not attain this depth. We also use continuous variables $y_{tv} \in [0,1]$ for each $v \in \cV^I_t \cup \cV^L_t$ to represent the flow of the counterfactual example in the decision tree. These variables are not explicitly defined as binary but will effectively take value $1$ if the counterfactual example passes through vertex $v$, and $0$ otherwise. This behavior is ensured with the following set of constraints:
\begin{align}
& y_{t1} = 1  & t \in \cT \label{branch:begin} \\
& y_{tv} = y_{tl(v)} + y_{tr(v)} & t \in \cT, \smash{v \in \cV^I_t} \\
& \sum_{v \in \cV^I_{td}} y_{tl(v)} \leq \lambda_{td} & t \in \cT,  \smash{d \in \cD_t} \tag{10a}  \\
& \sum_{v \in \cV^I_{td}} y_{tr(v)} \leq 1-\lambda_{td} & t \in \cT,  \smash{d \in \cD_t} \tag{10b}  \\
\setcounter{equation}{10}
& y_{tv} \in [0,1]  & t \in \cT, \smash{v \in \cV^I_t \cup \cV^I_t} \\
& \lambda_{td} \in \{0,1\} & t \in \cT, \smash{d \in \cD_t}.
\label{branch:end}
\end{align}

\begin{theorem}
\emph{
Formulation (\ref{branch:begin}--\ref{branch:end}) guarantees the integrality of the $\mathbf{y}$ variables. 
}
\label{t1}
\end{theorem}

Proof of this theorem is provided in the supplementary material. From these variables, finding the desired counterfactual class through majority vote can be expressed as:
\begin{align}
& z_{c} = \sum_{t \in \cT} \sum_{v \in \cV^L_t} w_{t} p_{tvc} y_{tv} & c \in \cC \label{model:classes}\\
& z_{c^*} > z_{c} & c \in \cC, c \neq c^*. \label{model:counterfact}
\end{align}

\subsection{Feature Consistency with the Splits}
\label{sec:model2}

The previous variables and constraints define the counterfactual example's paths through each tree. Additional constraints are needed to ensure that there exist feature values that are consistent with all these branching decisions. To that extent, we propose efficient formulations for each main data type (numerical, binary, and categorical), which can be combined in the case of heterogeneous data sets. We will refer to $I_\textsc{N}$, $I_\textsc{B}$, and $I_\textsc{C}$ as the index sets of each feature type.

\textbf{Numerical Features.} 
We can assume, w.l.o.g., that continuous features have been scaled into the interval $[0,1]$. For each numerical (continuous or discrete) feature~$i \in I_\textsc{N}$,
let~$k_i$ be the overall number of distinct split levels in the forest. Moreover, let $\smash{x^j_i}$ be the $\smash{j^\text{th}}$ split level for $j \in \{1,\dots,k_i\}$, and define $x^0_i = 0$ as well as $\smash{x^{k_i+1}_i = 1}$.

For each tree $t$, let $\smash{\cV^I_{tij}}$ be the set of internal nodes involving a split on feature~$i$ with level~$\smash{x^j_i}$, such that samples with feature values $x_i \leq \smash{x^j_i}$ descend to the left branch, whereas others values satisfying $x_i > \smash{x^j_i}$ descend to the right branch.
With these definitions, the consistency of a feature~$i$ through the forest can be modeled with the help of auxiliary continuous variables $\mu_i^{j}$ for $j \in \{0,\dots,k_i\}$, constrained in such a way that $\smash{\mu_i^{j}} = 0$ implies $x_i \in [0,\smash{x^j_i}]$ and
$\smash{\mu_i^{j}} = 1$ implies $x_i \in [\smash{x^j_i} + \epsilon ,1]$, where $\epsilon$ is a small constant.
These conditions are ensured through the following set of constraints:
\begin{align}
& \mu_i^{j-1} \geq \mu_i^j & j \in \{1,\dots,k_i\} \label{osimplex:begin} \\
& \mu_i^j \leq 1-y_{tl(v)} & j \in \{1,\dots,k_i\}, t \in \cT, \smash{v \in \cV^I_{tij}} \\
& \mu_i^{j-1} \geq y_{tr(v)} & j \in \{1,\dots,k_i\}, t \in \cT, \smash{v \in \cV^I_{tij}}  \\
& \mu_i^{j} \geq \epsilon y_{tr(v)} & j \in \{1,\dots,k_i\}, t \in \cT, \smash{v \in \cV^I_{tij}}  \\
& \mu_i^{j} \in [0,1] & j \in \{0,\dots,k_i\},
\label{osimplex:end}
\end{align}
and the feature level $x_i$ can be derived from these auxiliary variables (if
needed) as:
\begin{align}
x_i = \sum_{j = 0}^{k_i} (x^{j+1}_i - x^{j}_i) \mu_i^j.
\label{form-xi}
\end{align}

\textbf{Binary Features.} We assume, w.l.o.g., that all splits on binary features send values~$0$ to the left branch and values~$1$ to the right branch. Let $\cV^I_{ti}$ be the set of all vertices splitting on a binary feature $i \in I_\textsc{B}$. The consistency of this feature value can be ensured as follows:
\begin{align}
& x_i \leq \smash{1-y_{tl(v)}} & \smash{t \in \cT, v \in \cV^I_{ti}} \\
& x_i \geq y_{tr(v)} & \smash{t \in \cT, v \in \cV^I_{ti}} \\
& x_i \in \{0,1\}.\label{dom2}
\end{align}

\noindent
\textbf{Categorical Features.} Let~$k_i$ be the number of possible categories for feature $i \in I_\textsc{C}$. Let $\nu^j_i$ be a variable that will take value $1$ if~$x_i$ belongs to category~$j \in \{1,\dots, k_i\}$ and $0$ otherwise. Decision trees usually handle categorical variables through one-vs-all splits, sending the samples of a given category $j$ to the right branch and the rest of the samples to the left. Let $v \in \smash{\cV^I_{tij}}$ be the set of vertices splitting category $j$ of feature $i$. The consistency of this feature through the forest is modeled as:
\begin{align}
& \nu^j_i \leq 1-y_{tl(v)} & j \in C_i , t \in \cT, \smash{v \in \cV^I_{tij}} \label{categ:begin} \\
& \nu^j_i \geq y_{tr(v)} & j \in C_i, t \in \cT, \smash{v \in \cV^I_{tij}} \label{categ:begin2}  \\
& \nu^j_i  \in \{0,1\} &  j \in C_i \label{categ:begin3}  \\
& \sum_{j \in C_i} \nu^j_i = 1. \label{categ:end}
\end{align}

\begin{theorem}
\emph{
Formulation (\ref{branch:begin}--\ref{categ:end})
\vspace*{-0.2cm}
\begin{itemize}[nosep]
\item[(i)] yields feature values that are consistent with all splits in the forest;
\item[(ii)] involves only $\cO(N_v)$ non-zero terms in the constraint matrix overall;
\item[(iii)] achieves an equal or tighter linear relaxation than the decision logic constraints used in \citet{Cui2015}.
\end{itemize}
}
\label{t2}
\end{theorem}

Using a reduced number of integer variables is usually beneficial for computational performance. As discussed in the supplementary material, as a consequence of the integrality of the $\bm{\lambda}$ and~$\mathbf{y}$ variables, the domain of the $\mathbf{x}$, $\bm{\nu}$ variables in Formulation~\mbox{(\ref{branch:begin}--\ref{categ:end})} can be relaxed to the continuous interval $[0,1]$ while retaining integrality of the linear-relaxation solutions. We also show in the supplementary material how to efficiently handle ordinal features (or, generally, any ordered feature in which the open intervals between successive levels bear no meaning), multivariate splits on numerical features \citep{Brodley1995}, and combinatorial splits involving several categories for categorical features.

\subsection{Objective Function}
\label{sec:model3}

\begin{table*}[!t]
\centering
\caption{Domain knowledge and actionability constraints}
\label{tab:actionability}
\setlength{\tabcolsep}{0.3cm}
\scalebox{0.97}
{
\begin{tabular}{lr}
\toprule
\textbf{Domain Knowledge} & \textbf{Constraints} \\
\midrule
Fixed features & $x_i = \hat{x}_i$, $\mu_i = \hat{\mu}_i$, $\nu_i = \hat{\nu}_i$  \\
Monotonic features & $x_i \geq \hat{x}_i$, $\mu_i \geq \hat{\mu}_i$,  $\nu_i \geq \hat{\nu}_i$ \\
\midrule
Known \textbf{linear relations} between features & \multirow{2}{*}{$A(x_i - \hat{x}_i) \leq \mathbf{b}$} \\
(i.e., joint actionability
-- \citealt{Venkatasubramanian2020})\\
\midrule
Known \textbf{logical implications} between features, \\
Example for binary features $(x_1 = \textsc{True}) \Rightarrow (x_2$ = \textsc{True}) & $x_2 \geq x_1$ \\
Example for categorical features $x_1 \in \{\textsc{Cat1},\textsc{Cat2}\} \Rightarrow x_2 \in \{\textsc{Cat3},\textsc{Cat4}\}$ & $\nu^3_2 + \nu^4_2 \geq \nu^1_1 + \nu^2_1$ \\
\midrule
Resource constraints (e.g., time) as modeled by additional functions $g_i(\mathbf{x},\bm{\nu},\bm{\mu})$& $g_i(\mathbf{x},\bm{\nu},\bm{\mu}) \leq b_i$ \\
\bottomrule
\end{tabular}%
}
\end{table*}%

In a similar fashion as Gower's distance \citep{Gower1971}, we use a general objective  $f(\mathbf{x},\bm{\mu},\bm{\nu}) =
f^\textsc{n}(\bm{\mu}) +
f^\textsc{b}(\mathbf{x}) + f^\textsc{c}(\bm{\nu})$ which contains different terms to model the objective contributions of the numerical, binary, and categorical features. Moreover, as  actionability depends on each feature and direction of action, we will use asymmetric extensions of common distances metrics with feature-dependent weights.

The model presented in the previous sections gives a direct access to the values of the \textbf{binary} and \textbf{categorical} features through $\mathbf{x}$ and $\bm{\nu}$. We can therefore generally associate any discrete cost value for each of these choices:
\begin{align}
&f^\textsc{b}(\mathbf{x}) = \sum_{i \in I_\textsc{B}} ( c^\textsc{True}_i x_i +  c^\textsc{False}_i (1-x_i)) & \\
& f^\textsc{c}(\bm{\nu}) = \sum_{\smash{i \in I_\textsc{C}}} \sum_{\smash{j \in C_i}} c^j_i \nu^j_i. &
\end{align}
The cost coefficients corresponding to the origin state $\hat{\mathbf{x}}$ typically have zero cost, though negative values could be used if needed to model possible feature states that appear more desirable.

\textbf{Numerical} features can be directly accessed through the~$\mathbf{x}$ variables defined in Equation~(\ref{form-xi}), or indirectly through the~$\bm{\mu}$ variables. Most classical objectives used in counterfactual explanations can be directly expressed from $\mathbf{x}$, but modeling via~$\bm{\mu}$ can in some cases lead to a better linear relaxation (e.g., for $l_0$). To that end, we add the origin level~$\hat{x}_i$ (with index denoted as~$\hat{j}_i$) to the list of hyperplane levels defining the $\bm{\mu}_i$ variables. The $l_0$, $l_1$ and $l_2$ objectives with asymmetric and feature-dependent weights $\{c^-_i,c^+_i\}$ can then be expressed as:
\begin{align}
&l_0: \left\{
\begin{aligned}
& f_0^\textsc{n}(\bm{\mu}) = \sum_{i \in I_\textsc{N}} ( c^-_i z^-_i + c^+_i z^+_i) \hspace*{-0.3cm} \\
& z^-_i \geq 1-\mu^{j-1}_i, z^+_i \geq \mu^{j}_i & i \in I_\textsc{N}, j = \hat{j}_i  \\
& z^-_i \in \{0,1\}, z^+_i \in \{0,1\} & i \in I_\textsc{N}
\end{aligned}
\right.\label{l0}\\
& \nonumber \\
&l_1: \left\{
\begin{aligned}
&f_1^\textsc{n}(\bm{\mu}) = \sum_{j = 0}^{k_i}  (\phi^{j+1}_i - \phi^{j}_i) \mu_i^j & \\
& \text{with parameter } \phi^{j}_i = c^-_i \max(\hat{x}_i - x^{j}_i,0) \\
& \hspace*{2.62cm} + c^+_i \max(x^{j}_i - \hat{x}_i,0)
\end{aligned}
\right.\label{l1}\\
& \nonumber \\
&l_2: \left\{
\begin{aligned}
& f_2^\textsc{n}(\mathbf{x}) = \sum_{i \in I_\textsc{N}} \left( c^-_i (z^-_i)^2 + c^+_i (z^+_i)^2 \right) \hspace*{-0.3cm} \\
& z^+_i + z^-_i = x_i - \hat{x}_i & i \in I_\textsc{N} \\
& z^+_i \in \{0,1\}, \ z^-_i \in [0,1] & i \in I_\textsc{N}
\end{aligned}
\right.\label{l2}
\end{align}

Finally, observe that Equation~(\ref{l1}) can be extended to model any objective expressed as a piecewise-linear convex function by defining different $\bm{\phi}$ parameters and introducing extra $\bm{\mu}$ variables for any additional breakpoint needed.

Overall, our model gives an extensible framework for efficiently modeling most existing data types, decision-tree structures, and objectives. As it mathematically represents the space of all feasible counterfactual explanations, solving it to optimality using state-of-the-art MILP solvers for the objective of choice permits to locate optimal explanations.

\subsection{Domain Knowledge and Actionability}
\label{sec:model4}

As seen in the previous sections, Formulation~(\ref{branch:begin}--\ref{categ:end}) accounts for different data types (e.g., numerical, binary and categorical) without transformation. For categorical variables in particular, this effectively ensures that counterfactual explanations respect the structure of the data (select exactly one category). Our model's flexibility also permits us to integrate, as needed, additional domain knowledge and actionability requirements through linear and logical constraints. Table~\ref{tab:actionability} summarizes several of these constraints based on recent proposals from the literature. In the next section, we will detail our proposal to exploit isolation forests to ensure counterfactual explanations plausibility within our mathematical framework.

\subsection{Isolation Forests for Plausibility} 
\label{metho5}

We propose to rely on isolation forests \citep{Liu2008a} for a fine-grained representation of explanation plausibility. Isolation forests are trained to return an outlier score for any sample, inversely proportional to its average path depth within a set of randomized trees grown to full extent on random sample subsets. Therefore, constraining this average depth controls the outlier score (and consequently the plausibility) of the counterfactual explanation.

To include this constraint, we train the isolation forest $\cT_\textsc{I}$ on the training samples from the target class. Then, we mathematically express it through Formulation~(\ref{branch:begin}--\ref{branch:end}) and collect the $\bm{\mu}$ levels from the union of the two forests in Equations~\mbox{(\ref{osimplex:begin}--\ref{osimplex:end})}. Lastly, we constrain the average depth in $\cT_\textsc{I}$ as follows:
\begin{equation}
\sum_{t \in \cT_\textsc{I}} \sum_{v \in \cV^L_t} d_{tv} y_{tv} \geq \delta |\cT_\textsc{I}|,
\end{equation}
where $d_{tv}$ represents the depth of a vertex $v$ in tree $t$, and $\delta$ is a fixed threshold defining the average depth under which samples are declared as outliers. In our experiments, we will set this threshold to capture 10\% of the training data as an outlier, and therefore we seek a counterfactual explanation typical of the 90\% most common cases of the target class.

\section{Computational Experiments}
\label{sec:Experiments}

We conduct an extensive experimental campaign to fulfill two main goals:
\begin{itemize}
\item Evaluating the performance of our approach in terms of CPU time and solution quality. We measure the time needed to find optimal solutions on different data sets and evaluate the impact of the number of trees in the ensemble and their depth. We also compare the quality of our counterfactual explanations relatively to previous algorithms on a common objective ($l_1$ distance).
\item Assessing the impact of the \emph{plausibility} constraints obtained from isolation forests on the tractability of the model and the quality of the counterfactuals.
\end{itemize}

Our algorithm, referred to as OCEAN (Optimal Counterfactual Explanations) in the remainder of this paper, has been developed in Python 3.8 and can be readily executed from a single script that builds the most suitable mathematical model for the data set at hand. The complete data and source code needed to reproduce our experiments is provided at \url{https://github.com/vidalt/OCEAN}. The supplementary material of this paper also includes additional detailed results.

We use scikit-learn v0.23.0 for training random forests and Gurobi 9.1 (via gurobipy) for solving the mathematical models.
All experiments have been run on four threads of an Intel Core i9-9880H 2.30GHz CPU with 64GB of available RAM, running  Ubuntu 20.04.1 LTS. 

We now discuss the preparation of the data and describe each experiment.
We limit the scope of our experiments to the search of a single ---optimal--- explanation for each subject. If needed, diverse explanations could be generated by iteratively applying our framework, collecting its solution, and excluding it in subsequent iterations via an additional linear constraint or penalty term.

\subsection{Data Preparation}

We conduct our experiments on eight data sets representative of diverse applications such as loan approval, socio-economical studies, pretrial bail, news performance prediction, and malware detection. Table~\ref{tab:datasets} reports their number of samples ($n$), number of features (total~=~$p$, numerical~=~$p_\textsc{n}$, binary~=~$p_\textsc{b}$, and categorical~=~$p_\textsc{c}$), and source of origin.

\begin{table}[htbp]
\caption{Characteristics of the data sets}
\label{tab:datasets}
\setlength{\tabcolsep}{0.11cm}
\scalebox{0.9}
{
\begin{tabular}{lrrrrrr}
\toprule
\multicolumn{1}{l}{\textbf{Data set}} & \multicolumn{1}{r}{$n$} & \multicolumn{1}{r}{$p$} &
\multicolumn{1}{r}{$p_\textsc{n}$} &
\multicolumn{1}{r}{$p_\textsc{b}$} &
\multicolumn{1}{r}{$p_\textsc{c}$} &
\multicolumn{1}{r}{Src.} \\
\midrule
AD: Adult&45222&11&5&2&4&UCI\\
CC: Credit Card Default&29623&14&11&3&0&UCI\\
CP: COMPAS&5278&5&2&3&0&ProPublica\\
GC: German Credit&1000&9&5&1&3&UCI\\
ON: Online News &39644&47&43&2&2&UCI\\
PH: Data Phishing&11055&30&8&22&0&UCI\\
SP: Spambase&4601&57&57&0&0&UCI\\
ST: Students Performance&395&30&13&13&4&UCI\\
\bottomrule
\end{tabular}%
}
\end{table}%


All these data sets include heterogeneous feature types. For data sets AD, CC and CP, we used the same preprocessing as indicated in \citet{Karimi2020a}. GC preprocessing was done as suggested in \citet{UCIMachineLearning2017}. Finally, for ON we set ``data channel'' and ``weekday'' as categorical. The three remaining data sets are used in their original form. Each data set has been randomly split into 80\% training and 20\% test set.

To standardize the analyses between different data sets, we opted to set actionability constraints on two columns wherever applicable: ``age'' is constrained to be non-decreasing, and ``sex'' always stays fixed. 


\subsection{Performance and Scalability}

\begin{figure}[htbp]
\centering
\includegraphics[width=0.475\textwidth]{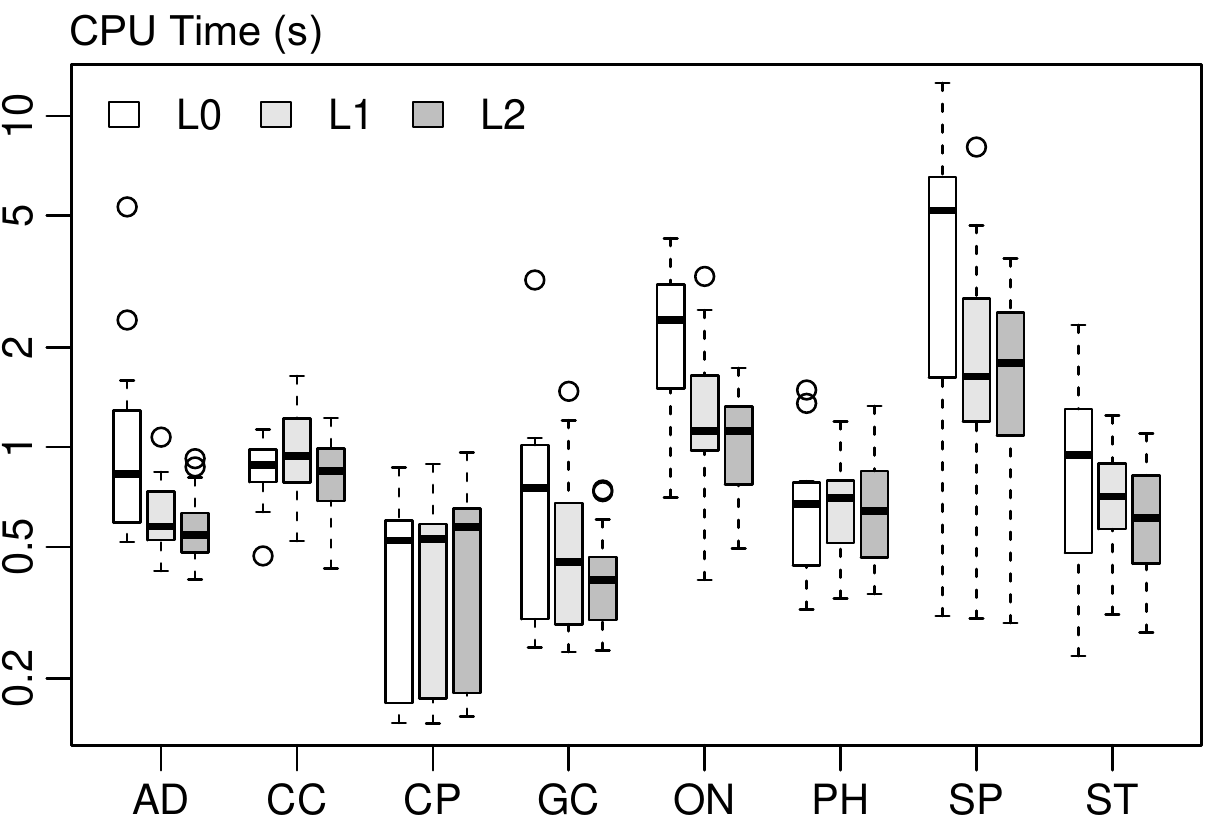}
\caption{CPU time to find an optimal counterfactual explanations, considering different data sets and objectives}
\label{fig:boxTime}
\end{figure}

In a first analysis, we evaluate the CPU time of OCEAN on each data sets for the asymmetric and weighted $l_0$, $l_1$, and $l_2$ objectives, \myblue{without the plausibility restrictions}. To simulate differences of actionability among features, the marginal weights~$c^-_i$~and~$c^+_i$ of each feature in the objective have been independently drawn in the uniform distribution $U(0.5,2)$. 
For each data set, we generated a single random forest with $100$ trees limited at depth $5$. We selected $20$ different negative samples from the test set to serve as origin points for the counterfactual explanations. Figure~\ref{fig:boxTime} reports the CPU time needed to find an optimal counterfactual explanation in each case. Each boxplot represents $20$ CPU time measurements, one for each counterfactual.

As visible in this experiment, OCEAN can locate \emph{optimal} counterfactual explanations in a matter of seconds, even for data sets including over fifty numerical features with a large number of levels. Finding counterfactual explanations with variants of the $l_1$ or~$l_2$ norms also appears slightly faster than with~$l_0$. For small data sets with a few dozen features, CPU times of the order of a second are typically achieved, making our framework applicable even in time-constrained environments, e.g., for interactive tasks. Finally, as counterfactual search has a decomposable geometrical structure, CPU time could be even further reduced by additional parallel computing if the need arises.

Next, we compare the performance of OCEAN with previous approaches: the heuristic feature tweaking (FT) algorithm of \citet{Tolomei2017a}, the exact model-agnostic counterfactual explanations (MACE) algorithm from \citet{Karimi2020a} and the optimal action extraction (OAE) approach proposed in \citet{Cui2015} and extended in \citet{Kanamori2020}. We used the implementation of FT and MACE (with precision $10^{-3}$) provided at \url{https://github.com/amirhk/mace}, and we provide a re-implementation of OAE within the same code base as OCEAN. For this comparative analysis, we use the $l_1$ objective with homogeneous weights, \myblue{as this is a common objective handled by all the considered methods. For a fair comparison, all the random forests and origin points for the counterfactual explanations have been saved in a serialized format, and identically loaded for each method.}

\begin{figure}[!h]
\centering
\includegraphics[width=0.48\textwidth]{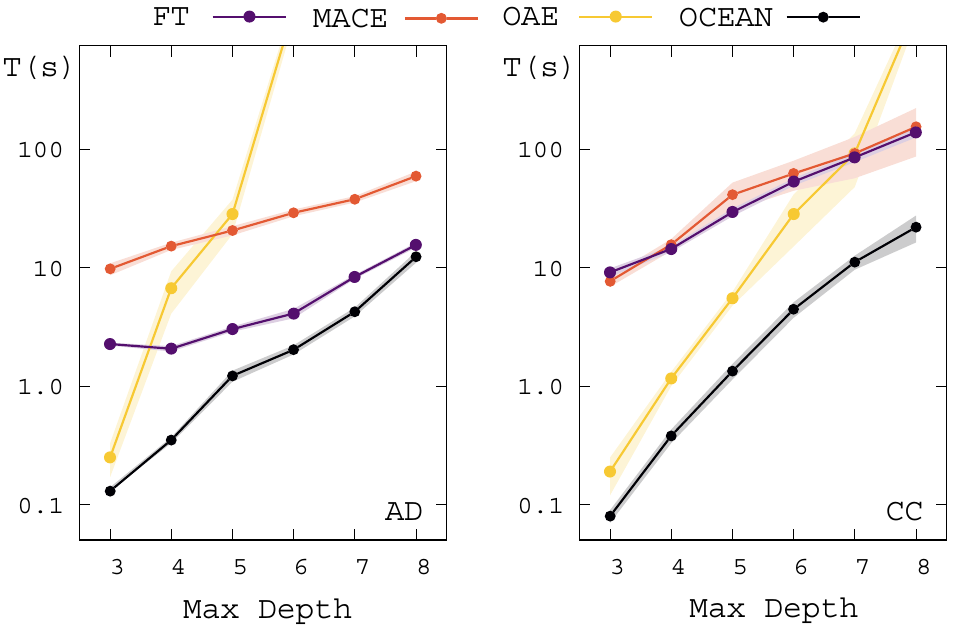}\vspace*{-0.4cm}
\caption{Comparative analysis of CPU time as a function of the maximum depth of the trees. Number of trees fixed to 100.}
\label{fig:lineDepth}
\end{figure}

\begin{figure}[!h]
\centering
\includegraphics[width=0.48\textwidth]{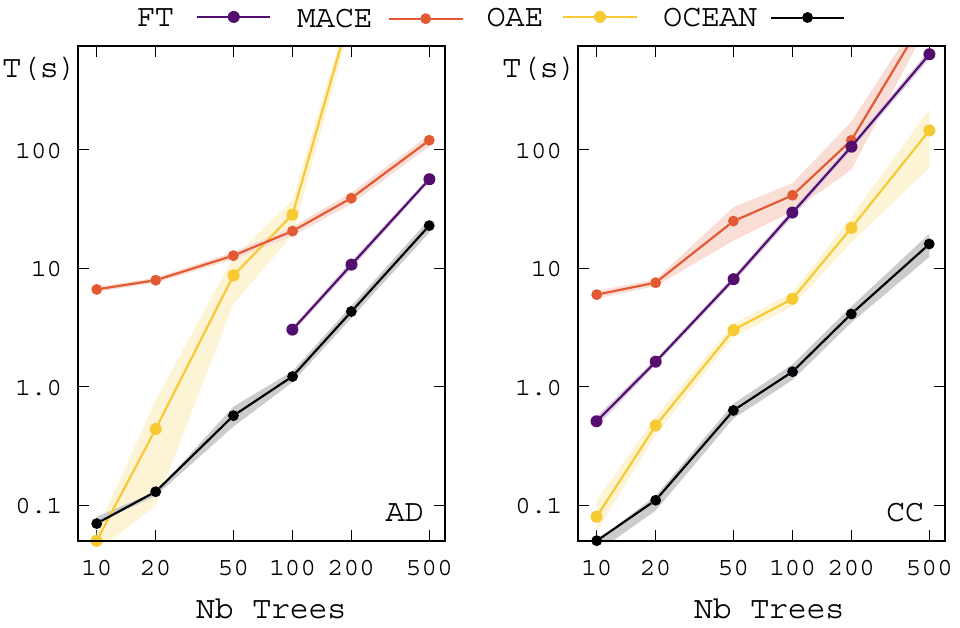}\vspace*{-0.4cm}
\caption{Comparative analysis of CPU time as a function of the number of trees in the ensemble. Maximum depth fixed to 5.}
\label{fig:lineTrees}
\end{figure}

We start from a baseline size of $100$ trees with a maximum depth of~$5$ for the random forest and then extend our analysis to a varying number of trees in $\{10,20,50,100,200,500\}$ and depth limit in $\{3,4,5,6,7,8\}$. 
Figures \ref{fig:lineDepth}~and~\ref{fig:lineTrees} report, for each method, the mean CPU time with its 95\% confidence interval as a function of these parameters for data sets AD~and~CC. \myblue{The same figures are provided in the supplementary material for the other data sets.}

As seen in these experiments, OCEAN locates optimal counterfactual explanations in a CPU time orders of magnitude smaller than MACE and OEA. Even in the most complex configurations (e.g., ST with $500$ trees), OCEAN completed the optimization within two minutes, whereas MACE and OEA ran for more than five hours without terminating.

\begin{table}[!h]
\caption{\myblue{Time and solution quality comparison}}
\label{res:time-quality}
\setlength{\tabcolsep}{0.15cm}
\scalebox{0.91}
{
\myblue{
\begin{tabular}{lc@{\hspace{0.2cm}}ccc@{\hspace{0.2cm}}ccc@{\hspace{0.2cm}}ccc@{\hspace{0.2cm}}c}
\toprule
\multicolumn{1}{l}{\textbf{Data}} & 
\multicolumn{2}{c}{\textbf{FT}} &&
\multicolumn{2}{c}{\textbf{MACE}} &&
\multicolumn{2}{c}{\textbf{OAE}} &&
\multicolumn{2}{c}{\textbf{OCEAN}} \\
&T(s)&R&&T(s)&R&&T(s)&R&&T(s)&R\\
\midrule
\textbf{AD}&3.03&15.9&&20.60&1.1&&28.37&1.0&&1.22&1.0\\
\textbf{CC}&29.44&10.2&&41.25&1.2&&5.52&1.0&&1.34&1.0\\
\textbf{CP}&22.68&4.5&&15.82&1.0&&0.38&1.0&&0.52&1.0\\
\textbf{GC}&16.26&4.8&&19.03&1.0&&5.08&1.0&&1.16&1.0\\
\textbf{ON}&10.05&31.7&&$>$900&---&&$>$900&---&&2.97&1.0\\
\textbf{PH}&10.95&1.4&&$>$900&---&&0.94&1.0&&0.52&1.0\\
\textbf{SP}&NA&---&&$>$900&---&&$>$900&---&&2.73&1.0\\
\textbf{ST}&NA&---&&$>$900&---&&69.64&1.0&&1.10&1.0\\
\bottomrule
\end{tabular}
}
}
\end{table}%

\myblue{Next, we evaluate the quality of the counterfactual explanations produced by these methods. We measure, for each method and dataset, the average CPU time per explanation and the ratio $R=D/D_\textsc{opt}$ between the sum of the $l_1$ distances of the $20$~counterfactual explanations produced by the method ($D$) and that of optimal explanations ($D_\textsc{opt}$). Table~\ref{res:time-quality} provides these values for the baseline setting ($100$ trees in the random forest limited at depth $5$), and similar results are provided in the supplementary material for the other settings. A value of ``NA'' means that no feasible counterfactual explanation has been found, whereas ``$>$900'' means that the CPU time limit of 900 seconds was exceeded.}

As observed in this experiment, it is notable that the CPU time of OCEAN is  shorter or comparable to that of the FT heuristic. Yet, despite its similar speed, FT does not guarantee optimality and effectively reached, on average, distance values that are \myblue{$1.4$ to $31.7$ times greater than the optima. It also regularly failed to find feasible counterfactual explanations (i.e., attaining the desired class) for SP and ST. These observations confirm the fact that heuristic techniques for counterfactual search can provide explanations that are considerably more complex than needed, and sensitive to the subject or subject group. In contrast, optimal counterfactual explanations are deterministic and fully specified through their mathematical definition ---independently of the search approach designed to find them--- providing us greater control and accountability. Arguably, ``optimal'' counterfactual explanations should be gradually established as a requirement for transparency and trustworthiness.}

Finally, the computational efficiency of OCEAN's can be explained, in part, by the sparsity of its mathematical model. For data set AD with baseline parameters, OEA's formulation included 163,605 non-zero terms, whereas OCEAN included only 18,330. This difference becomes even more marked as the maximum depth increases. With a maximum tree depth of $8$, the number of non-zero terms rises to 3,223,586 for OEA compared to 127,755 for OCEAN.
This permits integrating additional linear constraints, logical constraints, and a wide range of actionability definitions (Table~\ref{tab:actionability}). The next section will show how to profit from this performance gain to model fine-grained plausibility restrictions through isolation forests.

\subsection{Isolation Forests for Plausibility}
\label{exp:plausi}

\begin{figure}[!h]
\centering
\includegraphics[width=0.44\textwidth]{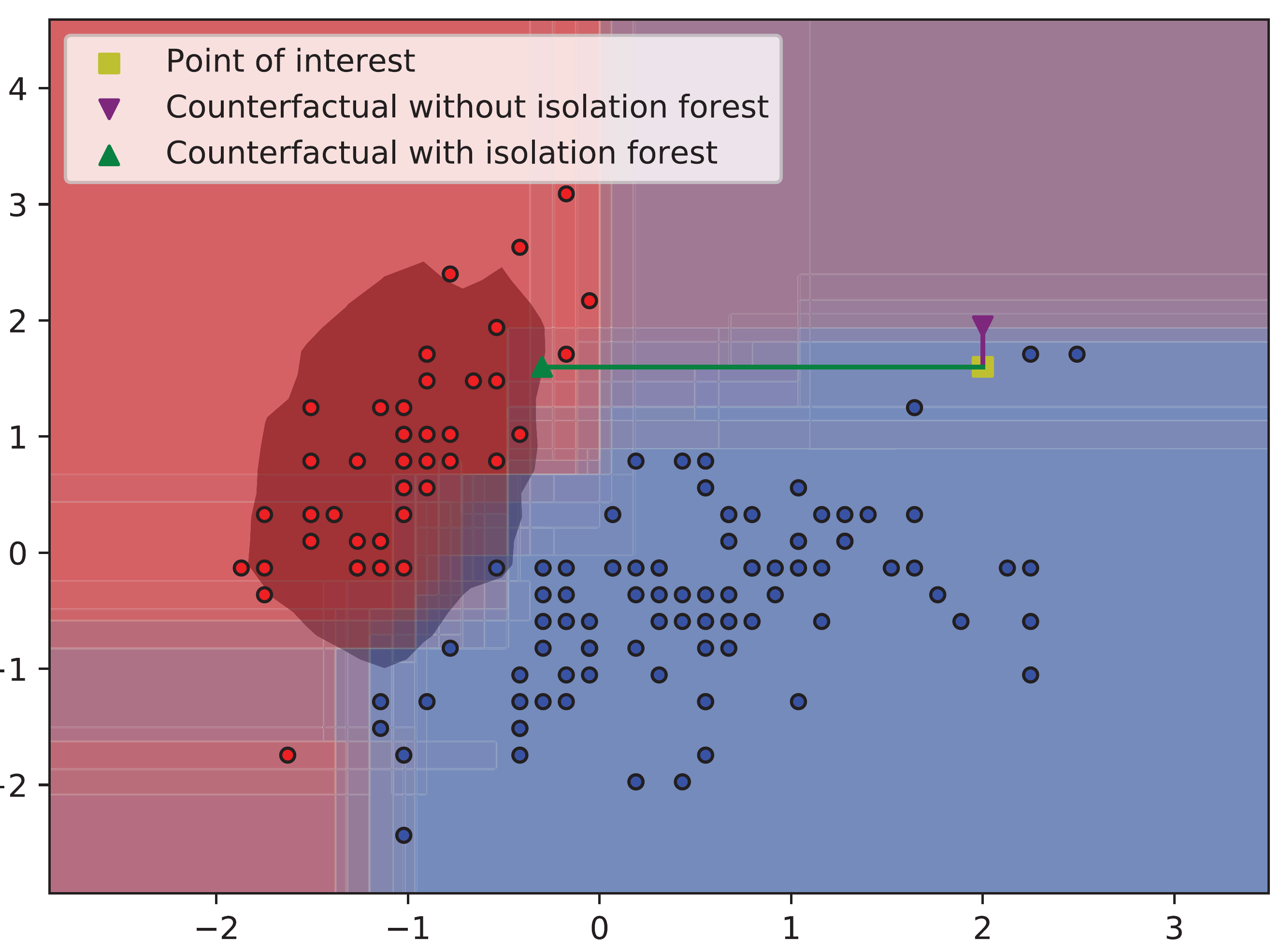}\vspace*{-0.1cm}
\caption{Isolation forests in counterfactual explanations for the Iris data set: x-axis = ``sepal length'', y-axis = ``sepal width''}
\label{fig:illus}
\end{figure}

We finally evaluate how isolation forests can ensure better plausibility in OCEAN's counterfactual explanations. Figure~\ref{fig:illus} first provides an illustrative example of our method on a small case. As observed in this example, random forests tend to make arbitrary class choices in low-density regions. Without any plausibility constraint, the counterfactual algorithm exploits the proximity to one such region to find a nearby counterfactual explanation. However, this explanation is not plausible as it represents an outlier among the samples of the desired~class. 

Introducing additional isolation forest constraints in our mathematical model as suggested in Section~\ref{metho5} permits us to circumvent this issue, as it successfully restricts the search for counterfactual examples to the core of the distribution of the desired class. Our restriction, therefore, builds on the same logic as the convex density constraints based on Gaussian mixtures proposed in \cite{Artelt2020}, though it has the general advantage to be distribution-agnostic and applicable to most data types.

To evaluate the impact of these plausibility constraints in our model, we conduct a final experiment which consists of measuring the prevalence of plausible explanations~(P), the cost of the explanations~(C), and the computational effort~(T) of OCEAN with and without isolation-forest constraints. The results of this analysis are reported in Table~\ref{res:plausibility}, with additional details in the supplementary material.

\begin{table}[htbp]
\caption{Impact of the plausibility constraints in OCEAN}
\label{res:plausibility}
\setlength{\tabcolsep}{0.23cm}
\scalebox{0.91}
{
\begin{tabular}{lccccccc}
\toprule
\multicolumn{1}{l}{\textbf{Data set}} & 
\multicolumn{3}{c}{\textbf{OCEAN-noIF}} &&
\multicolumn{3}{c}{\textbf{OCEAN-IF}} \\
&P&C&T(s)&&P&C&T(s)\\
\midrule
AD&55\%&0.21&0.75&&100\%&0.40&1.82\\
CC&0\%&0.03&0.99&&100\%&0.56&1.54\\
CP&25\%&0.12&0.44&&100\%&0.57&0.85\\
GC&25\%&0.09&0.60&&100\%&1.13&1.71\\
ON&100\%&0.01&1.34&&100\%&0.01&1.64\\
PH&35\%&0.78&0.36&&100\%&2.40&1.81\\
SP&100\%&0.02&2.71&&100\%&0.02&4.43\\
ST&40\%&1.18&0.62&&100\%&1.63&1.36\\
\bottomrule
\end{tabular}%
}
\end{table}%

As seen in this experiment, plausibility comes with extra costs that depend on the data set. Nevertheless, one cannot refer to a trade-off between cost and plausibility since misleading targets do not necessarily help to fulfill any given goal and may even trigger additional losses. Finding plausible and actionable explanations should therefore be the over-arching goal in counterfactual search. Our experiments also demonstrate that unconstrained solutions are rarely plausible and that dedicated constraints (e.g., through isolation forests) should be set up. Finally, the computational effort of OCEAN has roughly doubled due to the addition of the isolation forests in the model. This appears to be a reasonable increase given the current efficiency of OCEAN and the high practical importance of fine-grained plausibility constraints.

\section{Discussion}
\label{sec:Conclusions}

We have shown that it is possible to generate optimal plausible counterfactual explanations at scale for tree ensembles through careful integration of isolation forests and mathematical programming. By doing so, we have circumvented a major trustworthiness and accountability issue faced by heuristic approaches, and provided mathematical guarantees for tree ensembles interpretation. Our contribution also includes a modeling toolkit that can be used as a building block for a disciplined evaluation of counterfactual search models, plausibility constraints, and actionability paradigms in various applications.

The research perspectives are numerous. From a methodology standpoint, one can always challenge tractability limits and attempt to apply the OCEAN methodology to increasingly larger data sets and tree ensembles. To that end, we suggest to investigate new formulations and valid inequalities, as well as model compression and geometrical decomposition strategies \citep[see, e.g.,][]{Vidal2020a} which have the potential to speed up the solution process. We also recommend pursuing a disciplined evaluation of compression and explanation of white-box models through mathematical programming lenses, as performance guarantees are critical for a fair access to algorithmic recourse.

\section*{Acknowledgements}

\myblue{
This research has been partially supported by CAPES, CNPq [grant number 308528/2018-2] and FAPERJ [grant number E-26/202.790/2019] in Brazil.
}

\newpage


\onecolumn
\renewcommand{\thetable}{A\arabic{table}}
\renewcommand{\thefigure}{F\arabic{figure}}
\setcounter{figure}{0}
\setcounter{table}{0}

\section*{Supplementary Material -- Proofs}
\label{sec:proofs}

\noindent
\textbf{Proof of Theorem 1.}\\
Let $\bflambda, \bfy$ be a solution of Equations (8--12).
We will prove by induction on the depth $d$ that every node $v$ of depth $d$ in every tree $t \in \cT$ is such that the variables $y_{vt}$ belong to $\{0,1\}$ and that there is a unique node $v$ in $t$ at depth $d$ such that~$y_v=1$.
Equation~(8) gives the result for $d=1$.
Now, suppose that the result is true up to depth $d$ and let $v$ be a node of depth $d+1$ in some tree $t$ of the forest. 
Let $u$ be the parent of $v$.
By induction hypothesis, $y_u$ belongs to $\{0,1\}$.
If $y_u = 0$, then Equation~(9) implies that $y_v = 0$.
If $y_u = 1$, then $y_v = 1$ if $\lambda_{td} = 1$ (resp.~0) and $v$ is the left (resp.~right) child of $u$, and $0$ otherwise.
In all cases, we have $y_v \in \{0,1\}$.
Therefore, there is a unique node $v$ in $t$ of depth $d$ such that~$y_v=1$. This concludes the induction step and proves the theorem.

\noindent
\textbf{Proof of Theorem 2.}\vspace*{-0.3cm}
\begin{enumerate}
\item[(i)]
Let $\bflambda, \bfy, \bfx, \bfmu$ be a feasible solution. \\
We will first prove that the numeric features of this solution are consistent with all splits.\\
Theorem~1 ensures that $\bfy$ is integral and that in each tree $t$ there is a unique leaf $l$ such that $y_{l} = 1$. Let $P$ be the path from the root of $t$ to $l$. A backward induction along $P$ using Equation~(9) shows that $y_v = 1$ for every vertex~$v$ in $P$ and also implies that $y_v = 0$ for all nodes that are not in $P$.\\
Let $v$ be a non-leaf node along $P$. Let $i$ be the feature of~$v$ and $j$ be its split level (i.e., such that $v \in \smash{\calV_{tij}^I}$).
If~$P$ goes to the left child of $v$, then Equation~(16) ensures that $\smash{\mu_i^j} = 0$ and Equation~(15) ensures that $\smash{\mu_i^k} = 0$ for any~$k>j$.
Equation~(20) then ensures that $x_i \leq \smash{x_i^j}$.
Otherwise, if $P$ goes to the right child of $v$, then Equations~(17) and (15) ensure that $\mu_i^k = 1$ for any $k < j$. Equation~(18) ensures that $\mu_i^j > 0$, and Equation~(20) then gives $x_i > x_i^j$. Therefore, by disjunction of cases, $x_i$ is consistent with the split at node $v$.\\
If $i$ is a binary feature, then Equations (21--23) immediately ensure that $x_i = 0$ if $P$ goes to the left and $x_i = 1$ otherwise, which gives the consistency.
The same reasoning on the $\bm{\nu}$ gives the consistency for the categorical features.

\item[(ii)] 
This result immediately follows from the fact that the number of variables and constraints is in $\cO(N_v)$ and that the only constraints with more than two non-zero coefficients are constraints (10), (13), (20), and (27) with $\cO(N_v)$ non-zero terms overall in each case.

\item[(iii)]
We will use Cui et al.~(2015) notations when referring to the OAE formulation.
The proof reconstructs a solution of the linear relaxation of OAE from the optimal solution of the linear relaxation of OCEAN.
We prove this result when all the features are numeric.
The other cases are simpler and derive from a similar proof scheme.

Let $\bflambda, \bfy, \bfx, \bfmu$ be an optimal solution of the linear relaxation of our formulation.
We reconstruct a solution of the linear relaxation of OAE as follows.
First, remark that OAE's variables $\bfphi$ and OCEAN's variables $\bfy$ corresponding to tree leaves represent the same quantities. Therefore, given a tree $t$ and a leaf $k$ of $t$, define
$$ \phi_{tk} = y_{tk}$$ 
where the $\phi$ refer to the variables in OAE.
The flow constraints (8)-(9) of OCEAN then immediately imply
$$\sum_{k=1}^{m_t} \phi_{tk} = 1. $$

Let $i$ be a numerical feature and let $\hat{j}_i$ be the index associated with the original value $\hat x_i$ in OCEAN (see Section 3.3). 
OAE has the same splits as OCEAN except for the split level $\hat{j}_i$ corresponding to the original value of $\hat x_i$.
With this, OCEAN defines variables $\{\mu_{i}^{0},\dots,\mu_{i}^{k_i}\}$ and OAE defines variables $\{v_{i1},\dots,v_{in_i}\}$ with $n_i = k_i$. 


Now, let us set the following values:
\begin{equation}\tag{vars}\label{eq:vars}
v_{ij} = \tilde \mu_{i}^{j-1} - \tilde\mu_i^j
\quad
\text{ where }
\quad
\tilde{\mu}_{i}^{j} = 
\begin{cases}
\mu_{i}^{j} & \text{if } \mu_{i}^{j} > \epsilon \\
0 & \text{otherwise.} \\
\end{cases}
\end{equation}


Summing the values of $v_{ij}$ as defined in Equation~(\ref{eq:vars}), we obtain 
$\sum_j v_{ij} = \mu_i^0 - \tilde{\mu}_{i}^{k_i}$.
Moreover, observe that $\smash{\mu_i^0}$ does not intervene in constraints~(16) of OCEAN, and $\smash{\mu_i^{k_i}}$ does not intervene in constraints~(17).
The convexity of the cost and the optimality of the solution of OCEAN considered then imply that $\mu_i^{0} = 1$ and $\mu_i^{k_i} \leq \epsilon$ in an optimal solution of the linear relaxation, and therefore $\smash{\tilde{\mu}_i^{k_i} = 0}$, in such a way that:
$$\sum_{j=1}^{n_i} v_{ij} = 1.$$
Let us now prove that:
\begin{equation}\tag{dis}\label{eq:dis}
 \phi_{tk} \leq \sum_{v \in S_{kp}} v, \forall t, \forall k, \forall p \in \pi_{tk},
\end{equation}
which implies (by aggregation) the following constraints of OAE:
\begin{equation}\tag{agg}\label{eq:agg}
    \quad\phi_{tk} \leq \frac{1}{|\pi_{tk}|} \sum_{p \in \pi_{tk}} \sum_{v \in S_{kp}} v, \forall t, \forall k.
\end{equation}

For this, consider a tree $t$, a leaf $k$, and a node $p$ on the path to leaf $k$.
Let $j_p$ be the index of the split corresponding to~$p$.
Suppose that the path goes to the left at $p$.
Constraint (9) of OCEAN implies that \mbox{$\phi_{tk} = y_{tk} \leq y_{tl(p)}$}. 
Hence, $$\sum_{v \in S_{kp}} v = \mu_{i}^0 - \mu_{i}^{j_p-1} = 1 - \mu_{i}^{j_p-1},$$
and since $\smash{y_{tl(p)} \leq 1 - \mu_{i}^{j_p-1}}$ by
constraint~(15) and~(16) of OCEAN, we obtain~\eqref{eq:dis}.
Suppose now that the path goes to the right at $p$.
Constraint (9) of OCEAN implies that~\mbox{$\phi_{tk} = y_{tk} \leq y_{tr(p)}$}. 
Hence, $$\sum_{v \in S_{kp}} v = \mu_{i}^{j_p-1} - \mu_{i}^{n_i} = \mu_{i}^{j_p-1},$$ and since $\smash{y_{tr(p)} \leq \mu_{i}^{j_p-1}}$ by constraint~(17) of OCEAN, we obtain~\eqref{eq:dis}.

Moreover, the majority vote constraints (13) and (14) of OCEAN ensure that the majority vote constraint of OAE is satisfied, and thus the solution built is a feasible solution of OAE.

We finally evaluate the cost of this solution. 
We place ourselves in the general context of a piecewise linear convex loss function, as in Cui et al.~(2015). 
Define $\ell_i^j = \ell(\hat x_i, x_i^j)$, the value of the loss for feature $i$ if the counterfactual value for feature $i$ is $\smash{x_i^j}$, where $\smash{x_i^j}$ is the coordinate of split $j$ for feature $i$ in OCEAN and $1\leq j \leq n_i$.
Defining $\smash{x_i^0 = 0}$ and $x_i^{n_i+1} = 1$, the objective of OCEAN is:
$$\ell_i^0 + \sum_{j=0}^{n_i} (\ell_i^{j+1}-\ell_i^j)\mu_i^j. $$
Reindexing the sum and using the fact that $\tilde\mu_i^0 = 1$ and $\tilde\mu_i^{n_i} = 0$, we can rewrite this objective with an error non-greater than the numerical precision $\epsilon$ as:
$$ \sum_{j=1}^{n_i} \ell_i^j (\tilde\mu_i^{j-1} - \tilde\mu_i^j).$$
Let $b_i^j$ be the coordinate of the splits for OAE.
Finally, let $\hat j_i$ be the index of the split of OCEAN corresponding to $\hat x_i$.
We have $b_i^j = x_i^j$ for $j<\hat j_i$, and $b_i^j = x_i^{j+1}$ for $j\geq\hat j_i$.
Consider now the costs $C_{ij}(\bfx^c)$ in the objective of OAE, which correspond to $\ell(\hat x_i, \tilde x_i^j)$ where $\tilde x_i^j$ is the nearest point of $\hat x_i$ in the cell $j$ of OAE.
By convexity of the loss, $\tilde x_i^j$ is the right extremity of the cell if $j<\hat j_i$, the left extremity if $j>\hat j_i$ and $\hat x_i$ if $j=\hat j_i$.
Given the index shift between $x_i^j$ and $b_i^j$ in~$\hat x_i$, we get $C_{ij}(\bfx^c) = \ell(\hat x_i, x_i^j) = \ell_i^j$.
The value of the objective of OAE for the solution reconstructed is therefore
$$ \sum_{j=1}^{n_i}v_{ij}C_{ij} = \sum_{j=1}^{n_i}v_{ij}\ell_i^j = \sum_{j=1}^{n_i} \ell_i^j (\tilde\mu_i^{j-1} - \tilde\mu_i^j),$$
which is the value of the optimal solution of OCEAN.

In summary, we have built a feasible solution of the linear relaxation OAE whose objective is 
equal to the value of the linear relaxation of OCEAN. This concludes the proof.
\end{enumerate}

\noindent
\textbf{Implied integrality of the $\bm{x}$ variables for binary features and $\bm{\nu}$ variables for categorical features.}\\
Consider an optimal solution of the continuous relaxation of variables $\bm{x}$ for binary features and $\bm{\nu}$ for categorical features.
Fix all the variables but those corresponding to one of these features. 
Since the variables $y$ are integer in an optimal solution, the feasible set of the resulting problem is a simplex, and a basic optimal solution is therefore integer.

\section*{Supplementary Material -- Model Refinements}
\label{sec:ordinal}

In this section, we discuss some additional formulation refinement and extensions. Firstly, we provide improved formulations for ordinal features. Next, we discuss multivariate splits for numerical features and combinatorial splits for categorical features. 

\textbf{Ordinal features.} 
These features involve ordered levels in a similar fashion as numerical features, but open intervals between successive levels bear no meaning or should be prohibited. In this case, no fractional value should arise due to (e.g., actionability or plausibility) side constraints, and the associated costs are typically discretized over the ordinal levels. To efficiently model these features, we use specialized constraints that represent a simplification of the formulation used for numerical features.

Let~$k_i$ be the number of possible categories for ordinal feature $i \in I_\textsc{O}$. For each tree $t$, let $\smash{\cV^I_{tij}}$ be the set of internal nodes involving a split on feature~$i$ at level~$j$, such that samples with feature level smaller or equal to $j$ descend to the left branch, whereas other samples descend to the right branch. Consistency of feature~$i$ can be ensured through auxiliary variables $\omega_i^{j}$ for $j \in \{1,\dots,k_i-1\}$ which take value~$0$ if feature $i$ has a level smaller or equal to~$j$ and $1$ otherwise. These conditions can be ensured as follows:
\begin{align}
& \hspace*{-0.2cm}\omega_i^{j-1} \geq \omega_i^j & j \in \{2,\dots,k_i-1\}\hspace*{-0.1cm} \\
& \hspace*{-0.2cm}\omega_i^j \leq 1-y_{tl(v)}  \hspace*{-0.1cm} & j \in \{1,\dots,k_i \hspace*{-0.05cm}-\hspace*{-0.05cm} 1\}, t \in \cT, \smash{v \in \cV^I_{tij}}\hspace*{-0.1cm} \\
& \hspace*{-0.2cm}\omega_i^{j} \geq y_{tr(v)} & j \in \{1,\dots,k_i \hspace*{-0.05cm}-\hspace*{-0.05cm} 1\}, t \in \cT, \smash{v \in \cV^I_{tij}}\hspace*{-0.1cm}  \\
& \hspace*{-0.2cm}\omega_i^{j} \in \{0,1\} & j \in \{1,\dots,k_i-1\}.\hspace*{-0.1cm}
\end{align}

\textbf{Multivariate splits on numerical features.}
If the need arises, one can also model the search for counterfactual explanations in contexts where some splits of the decision trees (or isolation trees -- e.g., as in extended isolation forests) involve linear combinations of the features. To that end, we can rely on big-M constraints as follows:
\begin{align}
&\mathbf{a}_{tv} \mathbf{x} \leq b_{tv} - \epsilon   + M^+_{tv}(1-y_{tL(v)}) & t \in \cT, \smash{v \in \cV^I_t} \\
&\mathbf{a}_{tv} \mathbf{x} \geq b_{tv} + \epsilon - M^-_{tv} (1-y_{tR(v)}) & t \in \cT, \smash{v \in \cV^I_t}.
\end{align}
It is important to use the smallest possible value for the big-M constants to achieve a good linear relaxation. Given that numerical features are normalized in the interval $[0,1]$, we can use:
\begin{align}
M^+_{tv} &= \sum_{k=1}^p \max\{0,a_{tvp}\} - b_{tv} \\ M^-_{tv} &= b_{tv}- \sum_{k=1}^p \min\{0,a_{tvp}\}.
\end{align}

\textbf{Combinatorial splits on categorical features.} 
As usual for categorical features, $k_i$ will be used to denote the number of possible categories for feature $i \in I_\textsc{C}$, and let $\nu^j_i$ be a variable that will take value $1$ if~$x_i$ belongs to category~$j \in \{1,\dots, k_i\}$ and $0$ otherwise. Combinatorial splits on categorical features involve sending certain categories towards the right branch, and the rest on the left. For each split at an internal node $v$ of tree $t$, let $C^+_{ivt}$ be the set of categories that descend towards the right branch. Then, the consistency of categorical feature $i$ through the forest with combinatorial splits can be modeled as:
\begin{align}
& \sum_{j \in C^+_{ivt}} \nu^j_i \leq 1-y_{tl(v)} & t \in \cT, \smash{v \in \cV^I_{t}} \\
& \sum_{j \in C^+_{ivt}} \nu^j_i \geq y_{tr(v)} & t \in \cT, \smash{v \in \cV^I_{t}} \\
& \nu^j_i  \in \{0,1\} &  j \in C_i \\
& \sum_{j \in C_i} \nu^j_i = 1.
\end{align}





\section*{Supplementary Material -- Detailed Numerical Results}
\label{sec:detailed-results}

\subsection{Detailed Performance Comparisons}

\begin{figure}[!p]
\centering
\includegraphics[width=0.8\textwidth]{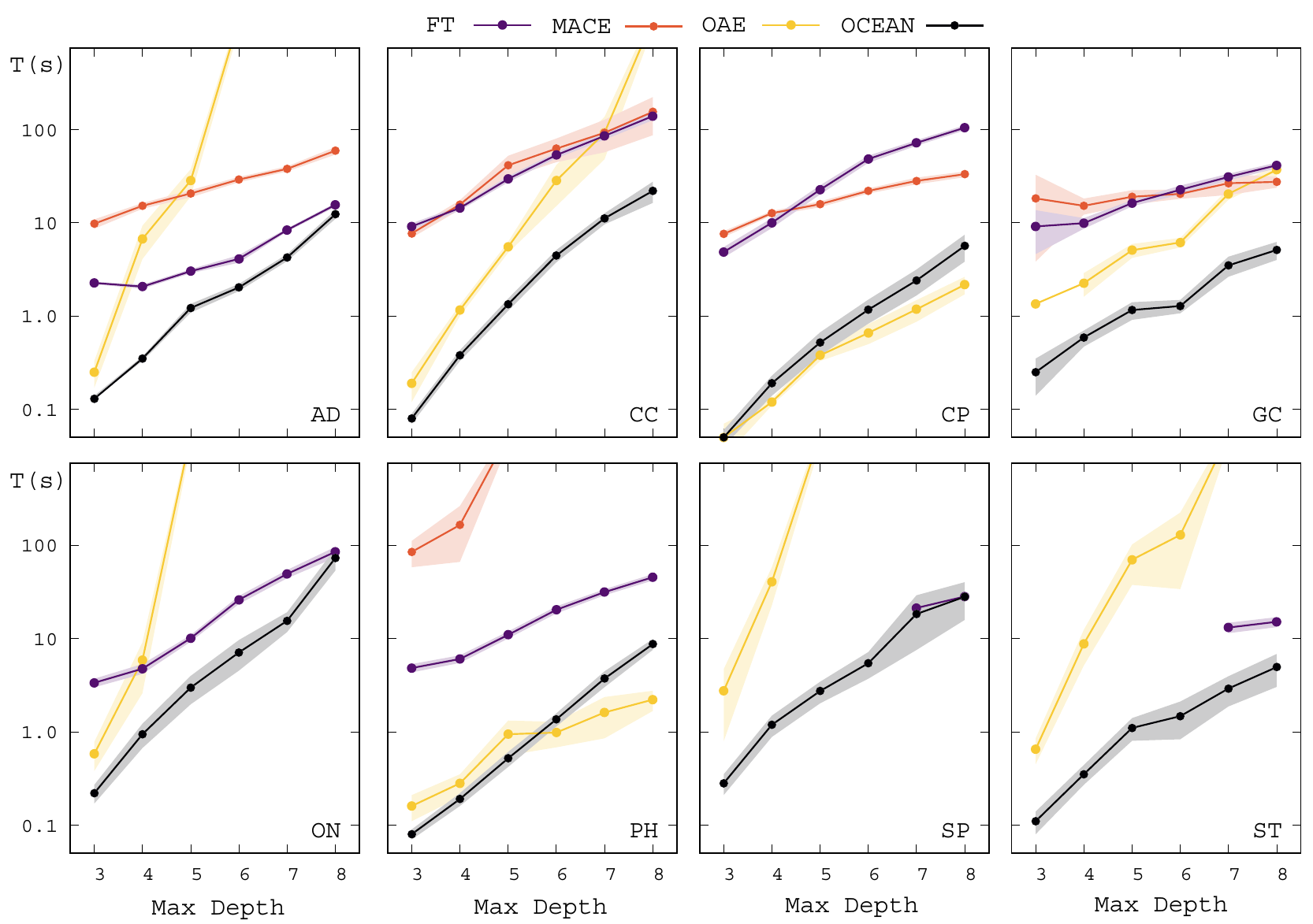}
\caption{Comparative analysis of CPU time as a function of the maximum depth of the trees. Number of trees set to 100. Results provided for all data sets.}
\label{detailedDepth}
\end{figure}

\begin{figure}[!p]
\centering
\includegraphics[width=0.8\textwidth]{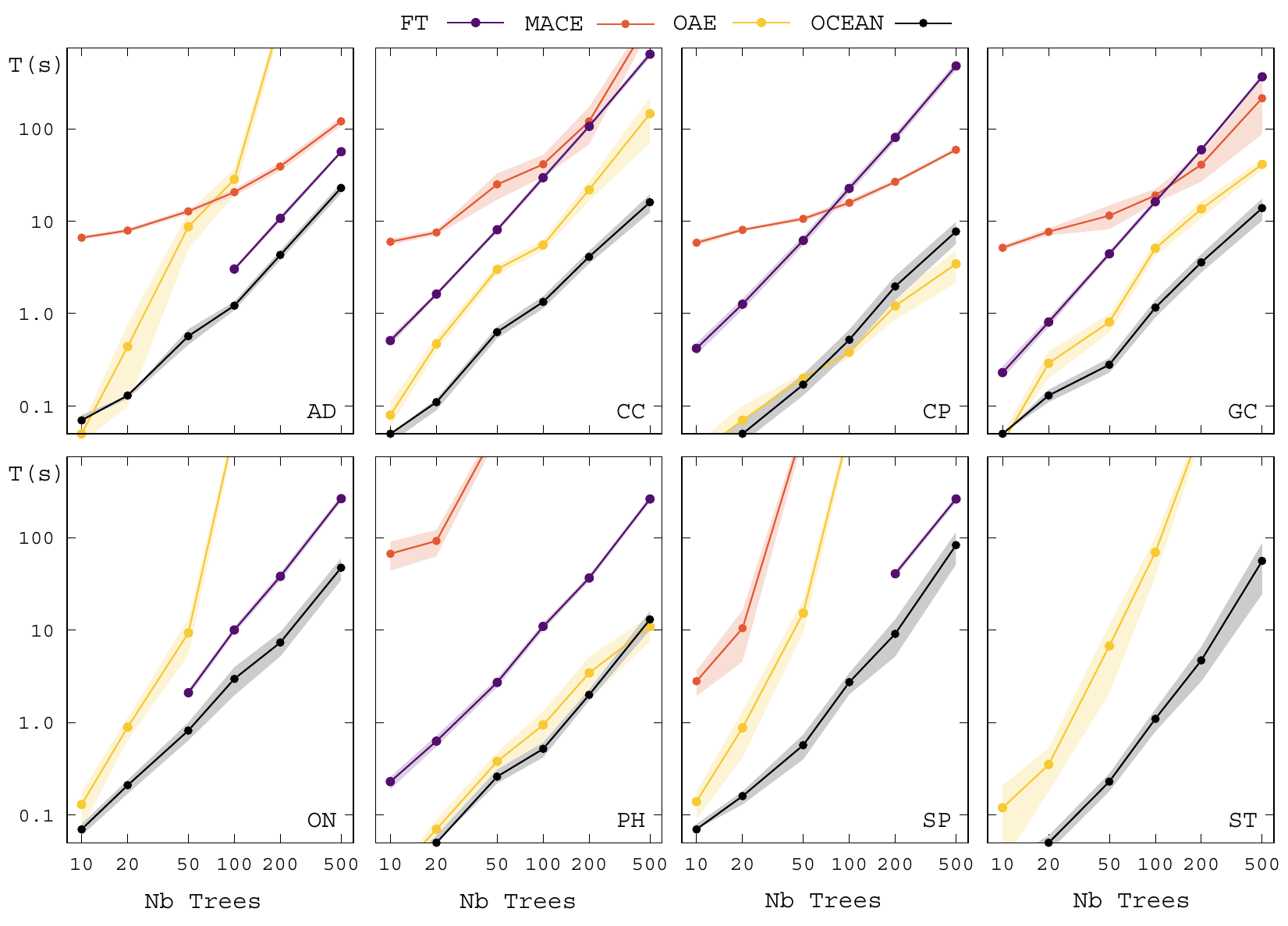}
\caption{Comparative analysis of CPU time as a function of the number of trees in the ensemble. Maximum depth fixed to 5. Results provided for all data sets.}
\label{detailedTrees}
\end{figure}

\myblue{
Figures~\ref{detailedDepth}~and~\ref{detailedTrees} extends Figures~2~and~3 in the main body of the paper with an analysis on all datasets and methods. In those figures, and missing data point for a method means that no feasible counterfactual has been found (possible for FT due to the way the search process is conducted) or that the CPU time limit of 900 seconds has been exceeded.
}

\myblue{
The left section of Table~\ref{detailed-results} extends the results of Table~3 in the main paper for a varying number of trees ``\#T'' in $\{10, 20, 50, 100, 200, \allowbreak 500\}$ with a maximum depth fixed to $5$. In a similar fashion, the right section of the table extends these results for varying depth ``\#D'' in $\{3,4,5,6,7,8\}$ with a number of trees fixed to $100$. 
}

\begin{table}[htbp]
\caption{Time and solution quality comparison, considering all configurations}
\label{detailed-results}
\setlength{\tabcolsep}{0.15cm}
\scalebox{0.8}
{
\myblue{
\begin{tabular}{
llc@{\hspace{0.2cm}}ccc@{\hspace{0.2cm}}ccc@{\hspace{0.2cm}}ccc@{\hspace{0.2cm}}c
c
llc@{\hspace{0.2cm}}ccc@{\hspace{0.2cm}}ccc@{\hspace{0.2cm}}ccc@{\hspace{0.2cm}}c
}
\toprule
\multicolumn{1}{l}{\textbf{\#T}} & \multicolumn{1}{l}{\textbf{Data}} & 
\multicolumn{2}{c}{\textbf{FT}} &&
\multicolumn{2}{c}{\textbf{MACE}} &&
\multicolumn{2}{c}{\textbf{OAE}} &&
\multicolumn{2}{c}{\textbf{OCEAN}} & \hspace*{0.5cm} &
\multicolumn{1}{l}{\textbf{\#D}} & \multicolumn{1}{l}{\textbf{Data}} & 
\multicolumn{2}{c}{\textbf{FT}} &&
\multicolumn{2}{c}{\textbf{MACE}} &&
\multicolumn{2}{c}{\textbf{OAE}} &&
\multicolumn{2}{c}{\textbf{OCEAN}} \\
&&T(s)&R&&T(s)&R&&T(s)&R&&T(s)&R&&&&T(s)&R&&T(s)&R&&T(s)&R&&T(s)&R\\
\midrule
\textbf{10}&\textbf{AD}&NA&---&&6.64&1.0&&0.05&1.0&&0.07&1.0&&\textbf{3}&\textbf{AD}&2.27&22.6&&9.76&1.0&&0.25&1.0&&0.13&1.0\\
&\textbf{CC}&0.51&23.0&&5.99&1.3&&0.08&1.0&&0.05&1.0&&&\textbf{CC}&9.13&128.2&&7.69&6.1&&0.19&1.0&&0.08&1.0\\
&\textbf{CP}&0.42&18.5&&5.85&1.0&&0.03&1.0&&0.02&1.0&&&\textbf{CP}&4.86&5.9&&7.59&1.0&&0.05&1.0&&0.05&1.0\\
&\textbf{GC}&0.23&17.4&&5.17&1.0&&0.04&1.0&&0.05&1.0&&&\textbf{GC}&9.14&1.9&&18.27&1.0&&1.35&1.0&&0.25&1.0\\
&\textbf{ON}&0.13&106.0&&$>$900&---&&0.13&1.0&&0.07&1.0&&&\textbf{ON}&3.35&38.5&&$>$900&---&&0.58&1.0&&0.22&1.0\\
&\textbf{PH}&0.23&1.7&&67.13&1.0&&0.02&1.0&&0.02&1.0&&&\textbf{PH}&4.82&1.3&&84.46&1.0&&0.16&1.0&&0.08&1.0\\
&\textbf{SP}&NA&---&&2.80&7.6&&0.14&1.0&&0.07&1.0&&&\textbf{SP}&NA&---&&$>$900&---&&2.74&1.0&&0.28&1.0\\
&\textbf{ST}&NA&---&&$>$900&---&&0.12&1.0&&0.02&1.0&&&\textbf{ST}&NA&---&&$>$900&---&&0.65&1.0&&0.11&1.0\\
\midrule
\textbf{20}&\textbf{AD}&NA&---&&7.92&1.0&&0.44&1.0&&0.13&1.0&&\textbf{4}&\textbf{AD}&2.07&22.5&&15.19&1.0&&6.71&1.0&&0.35&1.0\\
&\textbf{CC}&1.62&22.2&&7.56&1.3&&0.47&1.0&&0.11&1.0&&&\textbf{CC}&14.40&26.5&&15.66&1.4&&1.16&1.0&&0.38&1.0\\
&\textbf{CP}&1.27&10.6&&8.07&1.0&&0.07&1.0&&0.05&1.0&&&\textbf{CP}&10.00&4.9&&12.68&1.0&&0.12&1.0&&0.19&1.0\\
&\textbf{GC}&0.81&10.0&&7.69&1.0&&0.29&1.0&&0.13&1.0&&&\textbf{GC}&9.92&5.2&&15.21&1.0&&2.25&1.0&&0.59&1.0\\
&\textbf{ON}&NA&---&&$>$900&---&&0.89&1.0&&0.21&1.0&&&\textbf{ON}&4.74&52.1&&$>$900&---&&5.86&1.0&&0.94&1.0\\
&\textbf{PH}&0.63&1.4&&92.15&1.0&&0.07&1.0&&0.05&1.0&&&\textbf{PH}&6.02&1.5&&164.22&1.0&&0.28&1.0&&0.19&1.0\\
&\textbf{SP}&NA&---&&10.51&4.9&&0.88&1.0&&0.16&1.0&&&\textbf{SP}&NA&---&&$>$900&---&&40.31&1.0&&1.19&1.0\\
&\textbf{ST}&NA&---&&$>$900&---&&0.35&1.0&&0.05&1.0&&&\textbf{ST}&NA&---&&$>$900&---&&8.76&1.0&&0.35&1.0\\
\midrule
\textbf{50}&\textbf{AD}&NA&---&&12.80&1.1&&8.71&1.0&&0.57&1.0&&\textbf{5}&\textbf{AD}&3.03&15.9&&20.60&1.1&&28.37&1.0&&1.22&1.0\\
&\textbf{CC}&8.06&5.9&&24.97&1.1&&3.02&1.0&&0.63&1.0&&&\textbf{CC}&29.44&10.2&&41.25&1.2&&5.52&1.0&&1.34&1.0\\
&\textbf{CP}&6.19&5.0&&10.65&1.0&&0.20&1.0&&0.17&1.0&&&\textbf{CP}&22.68&4.5&&15.82&1.0&&0.38&1.0&&0.52&1.0\\
&\textbf{GC}&4.44&6.2&&11.51&1.0&&0.81&1.0&&0.28&1.0&&&\textbf{GC}&16.26&4.8&&19.03&1.0&&5.08&1.0&&1.16&1.0\\
&\textbf{ON}&2.10&97.7&&$>$900&---&&9.29&1.0&&0.82&1.0&&&\textbf{ON}&10.05&31.7&&$>$900&---&&$>$900&---&&2.97&1.0\\
&\textbf{PH}&2.72&1.6&&$>$900&---&&0.38&1.0&&0.26&1.0&&&\textbf{PH}&10.95&1.4&&$>$900&---&&0.94&1.0&&0.52&1.0\\
&\textbf{SP}&NA&---&&$>$900&---&&15.31&1.0&&0.57&1.0&&&\textbf{SP}&NA&---&&$>$900&---&&$>$900&---&&2.73&1.0\\
&\textbf{ST}&NA&---&&$>$900&---&&6.75&1.0&&0.23&1.0&&&\textbf{ST}&NA&---&&$>$900&---&&69.64&1.0&&1.10&1.0\\
\midrule
\textbf{100}&\textbf{AD}&3.03&15.9&&20.60&1.1&&28.37&1.0&&1.22&1.0&&\textbf{6}&\textbf{AD}&4.11&21.6&&29.08&1.1&&$>$900&---&&2.03&1.0\\
&\textbf{CC}&29.44&10.2&&41.25&1.2&&5.52&1.0&&1.34&1.0&&&\textbf{CC}&53.13&11.5&&62.36&1.2&&28.33&1.0&&4.45&1.0\\
&\textbf{CP}&22.68&4.5&&15.82&1.0&&0.38&1.0&&0.52&1.0&&&\textbf{CP}&48.20&9.1&&21.96&1.0&&0.66&1.0&&1.17&1.0\\
&\textbf{GC}&16.26&4.8&&19.03&1.0&&5.08&1.0&&1.16&1.0&&&\textbf{GC}&22.60&6.9&&20.41&1.0&&6.15&1.0&&1.28&1.0\\
&\textbf{ON}&10.05&31.7&&$>$900&---&&$>$900&---&&2.97&1.0&&&\textbf{ON}&25.94&24.7&&$>$900&---&&$>$900&---&&7.08&1.0\\
&\textbf{PH}&10.95&1.4&&$>$900&---&&0.94&1.0&&0.52&1.0&&&\textbf{PH}&20.30&1.3&&178.61&1.0&&0.98&1.0&&1.36&1.0\\
&\textbf{SP}&NA&---&&$>$900&---&&$>$900&---&&2.73&1.0&&&\textbf{SP}&NA&---&&$>$900&---&&$>$900&---&&5.42&1.0\\
&\textbf{ST}&NA&---&&$>$900&---&&69.64&1.0&&1.10&1.0&&&\textbf{ST}&NA&---&&$>$900&---&&128.52&1.0&&1.47&1.0\\
\midrule
\textbf{200}&\textbf{AD}&10.69&14.8&&39.18&1.0&&$>$900&---&&4.30&1.0&&\textbf{7}&\textbf{AD}&8.38&22.3&&37.84&1.1&&$>$900&---&&4.24&1.0\\
&\textbf{CC}&106.37&8.3&&120.31&1.2&&21.89&1.0&&4.12&1.0&&&\textbf{CC}&85.39&13.6&&92.11&1.2&&92.18&1.0&&11.13&1.0\\
&\textbf{CP}&80.53&3.4&&26.61&1.0&&1.21&1.0&&1.97&1.0&&&\textbf{CP}&71.92&8.2&&28.14&1.0&&1.18&1.0&&2.41&1.0\\
&\textbf{GC}&59.37&4.9&&40.91&1.0&&13.58&1.0&&3.59&1.0&&&\textbf{GC}&30.99&7.7&&26.50&1.0&&20.41&1.0&&3.49&1.0\\
&\textbf{ON}&38.18&18.4&&$>$900&---&&$>$900&---&&7.34&1.0&&&\textbf{ON}&49.16&70.9&&$>$900&---&&$>$900&---&&15.41&1.0\\
&\textbf{PH}&36.60&1.6&&$>$900&---&&3.44&1.0&&1.99&1.0&&&\textbf{PH}&31.33&1.3&&$>$900&---&&1.61&1.0&&3.72&1.0\\
&\textbf{SP}&40.71&8.5&&$>$900&---&&$>$900&---&&9.13&1.0&&&\textbf{SP}&21.22&7.8&&$>$900&---&&$>$900&---&&18.26&1.0\\
&\textbf{ST}&NA&---&&$>$900&---&&$>$900&---&&4.70&1.0&&&\textbf{ST}&13.07&3.3&&$>$900&---&&$>$900&---&&2.91&1.0\\
\midrule
\textbf{500}&\textbf{AD}&56.39&18.0&&120.25&1.1&&$>$900&---&&22.96&1.0&&\textbf{8}&\textbf{AD}&15.55&10.5&&59.30&1.1&&$>$900&---&&12.35&1.0\\
&\textbf{CC}&640.08&4.9&&$>$900&---&&145.36&1.0&&15.98&1.0&&&\textbf{CC}&139.40&18.2&&154.47&1.3&&$>$900&---&&21.96&1.0\\
&\textbf{CP}&479.78&2.9&&59.36&1.0&&3.45&1.0&&7.74&1.0&&&\textbf{CP}&104.49&9.1&&33.11&1.0&&2.17&1.0&&5.65&1.0\\
&\textbf{GC}&365.30&4.2&&214.64&1.0&&41.22&1.0&&13.82&1.0&&&\textbf{GC}&41.25&7.5&&27.45&1.0&&36.89&1.0&&5.11&1.0\\
&\textbf{ON}&265.22&33.0&&$>$900&---&&$>$900&---&&47.35&1.0&&&\textbf{ON}&84.72&60.3&&$>$900&---&&$>$900&---&&72.87&1.0\\
&\textbf{PH}&261.96&1.5&&$>$900&---&&10.96&1.0&&13.05&1.0&&&\textbf{PH}&45.32&1.3&&236.58&1.0&&2.21&1.0&&8.70&1.0\\
&\textbf{SP}&262.35&7.7&&$>$900&---&&$>$900&---&&83.05&1.0&&&\textbf{SP}&28.16&6.7&&$>$900&---&&$>$900&---&&27.94&1.0\\
&\textbf{ST}&NA&---&&$>$900&---&&$>$900&---&&55.95&1.0&&&\textbf{ST}&15.07&3.0&&$>$900&---&&$>$900&---&&4.93&1.0\\
\bottomrule
\end{tabular}
}
}
\end{table}%

\subsection{Detailed Performance with Plausibility Constraints}

This section provides additional detailed results concerning the performance of OCEAN with plausibility constraints via isolation forests.

\begin{figure}[htb]
\centering
\includegraphics[width=0.45\textwidth]{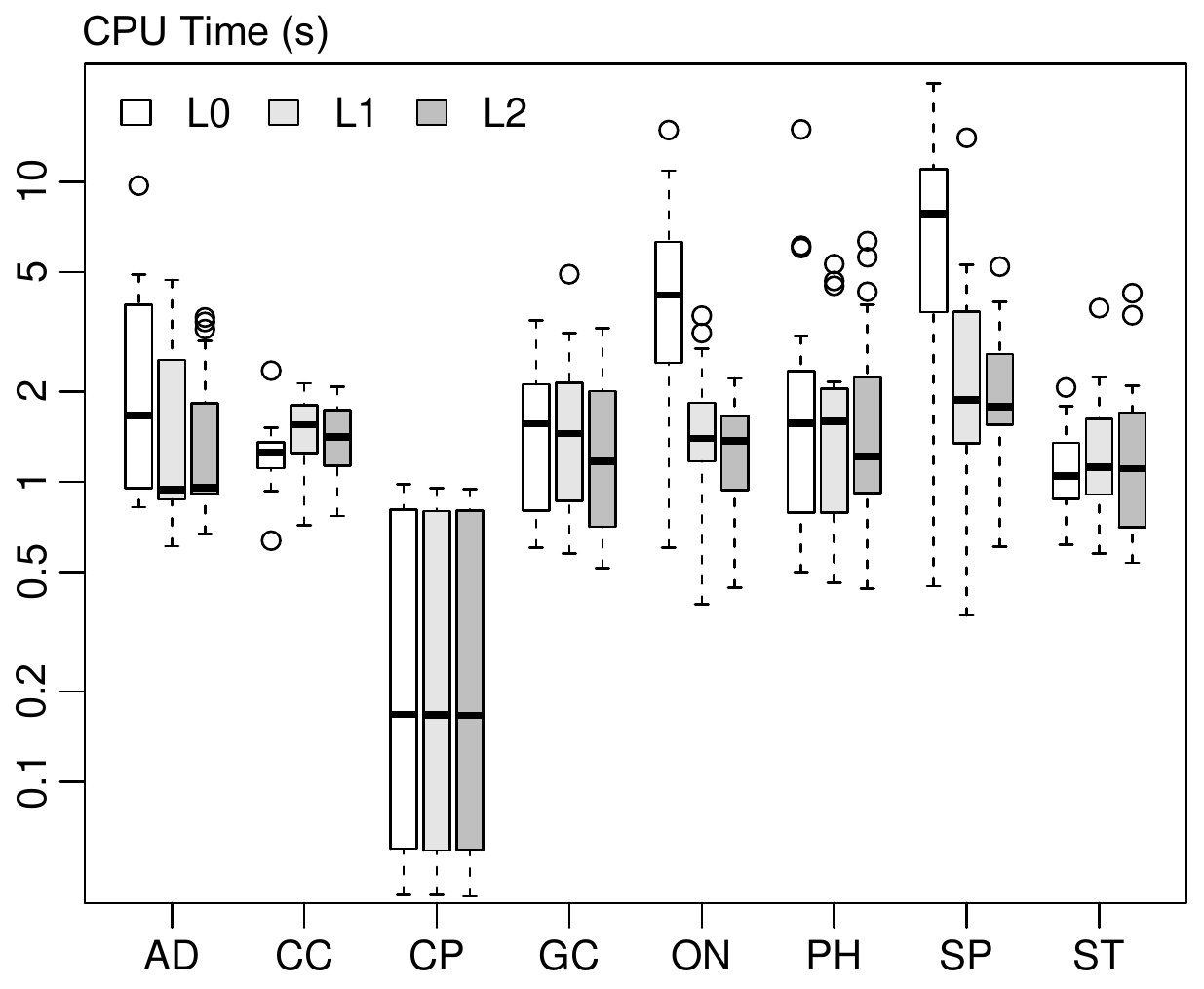}
\caption{CPU time to find an optimal counterfactual explanations, considering different data sets and objectives}
\label{fig:plausibilityNorms}
\end{figure}

Firstly, Figure~\ref{fig:plausibilityNorms} displays the same computational time analysis as Figure~1 in the main paper, when considering the additional plausibility constraints through isolation forests. As visible on this figure, OCEAN maintains a good scalability for all objectives even with the plausibility restrictions.

Finally, Table~\ref{res:plausibilityIF} reports the average computational time in seconds needed to find optimal counterfactual explanations with (OCEAN-IF) and without (OCEAN-noIF) plausibility constraints as a function of the maximum depth of the trees (in $\{5,6,7,8\}$, corresponding to a maximum of $\{32,64,128,256\}$ leaves). In these experiments, we use the $l_1$ objective, the number of trees is set to the baseline value of $100$, and the isolation forest has the same depth limit as the random forest.

\begin{table}[htb]
\centering
\caption{CPU time (s) of OCEAN with isolation forests for ensuring plausibility}
\label{res:plausibilityIF}
\setlength{\tabcolsep}{0.3cm}
\scalebox{0.85}
{
\begin{tabular}{lccccccccc}
\toprule
\multicolumn{1}{l}{\textbf{Data set}} & 
\multicolumn{4}{c}{\textbf{OCEAN-noIF}} &&
\multicolumn{4}{c}{\textbf{OCEAN-IF}} \\
Max-Depth&5&6&7&8&&5&6&7&8\\
Max-Leaves&32&64&128&256&&32&64&128&256\\
\midrule
AD&0.75&1.43&2.70&5.36&&1.82&3.83&7.26&14.10\\
CC&0.99&4.70&9.14&24.50&&1.54&5.42&12.83&32.83\\
CP&0.44&1.18&2.74&5.32&&0.85&1.35&3.18&7.06\\
GC&0.60&1.15&2.05&3.50&&1.71&3.59&11.04&18.16\\
ON&1.34&3.19&9.61&36.81&&1.64&4.29&11.79&41.87\\
PH&0.36&1.16&2.25&5.09&&1.81&5.86&15.16&43.12\\
SP&2.71&6.60&15.92&34.02&&4.43&7.18&27.63&37.98\\
ST&0.62&1.47&2.22&3.13&&1.36&3.63&8.13&12.31\\
\bottomrule
\end{tabular}%
}
\end{table}%

As visible in these experiments, the use of the plausibility restrictions through isolation forests roughly doubles the time needed to locate optimal explanations. This increase directly relates to the fact that considering both the random forest and isolation forest simultaneously involves considering twice the number of trees. Despite this increase of model complexity, optimal explanations are found in less than one minute, even when considering a maximum depth of $8$ (i.e., with up to $256$ leaves per tree and $51,200$ leaves overall in both forests). 

\section*{Supplementary Material -- Open Source Code}
\label{sec:source-code}

All the material (source code and data sets) \myblue{needed to reproduce our experiments is accessible at \url{https://github.com/vidalt/OCEAN} under a MIT license.}


\begin{thebibliography}{29}
\providecommand{\natexlab}[1]{#1}
\providecommand{\url}[1]{\texttt{#1}}
\expandafter\ifx\csname urlstyle\endcsname\relax
  \providecommand{\doi}[1]{doi: #1}\else
  \providecommand{\doi}{doi: \begingroup \urlstyle{rm}\Url}\fi

\bibitem[Artelt \& Hammer(2020)Artelt and Hammer]{Artelt2020}
Artelt, A. and Hammer, B.
\newblock {Convex density constraints for computing plausible counterfactual
  explanations}.
\newblock In \emph{International Conference on Artificial Neural Networks},
  pp.\  353--365. Springer Cham, 2020.

\bibitem[{Article 29 Data Protection Working Party}(2017)]{WorkingParty2018}
{Article 29 Data Protection Working Party}.
\newblock {Guidelines on Automated individual decision-making and profiling for
  the purposes of Regulation 2016/679}, 2017.

\bibitem[Barocas et~al.(2020)Barocas, Selbst, and Raghavan]{Barocas2020}
Barocas, S., Selbst, A., and Raghavan, M.
\newblock {The hidden assumptions behind counterfactual explanations and
  principal reasons}.
\newblock In \emph{Proceedings of the 2020 Conference on Fairness,
  Accountability, and Transparency}, pp.\  80--89. ACM, 2020.

\bibitem[Bixby(2012)]{Bixby2012}
Bixby, R.
\newblock {A brief history of linear and mixed-integer programming
  computation}.
\newblock \emph{Documenta Mathematica}, pp.\  107--121, 2012.

\bibitem[Breiman(2001)]{Breiman2001}
Breiman, L.
\newblock {Random forests}.
\newblock \emph{Machine Learning}, 45\penalty0 (1):\penalty0 5--32, 2001.

\bibitem[Brodley \& Utgoff(1995)Brodley and Utgoff]{Brodley1995}
Brodley, C. and Utgoff, P.
\newblock {Multivariate decision trees}.
\newblock \emph{Machine Learning}, 19\penalty0 (1):\penalty0 45--77, 1995.

\bibitem[Cui et~al.(2015)Cui, Chen, He, and Chen]{Cui2015}
Cui, Z., Chen, W., He, W., and Chen, Y.
\newblock {Optimal action extraction for random forests and boosted trees}.
\newblock \emph{Proceedings of the 21th ACM SIGKDD International Conference on
  Knowledge Discovery and Data Mining}, pp.\  179--188, 2015.

\bibitem[{German credit preprocessing}(2017)]{UCIMachineLearning2017}
{German credit preprocessing}, 2017.
\newblock URL \url{https://www.kaggle.com/uciml/german-credit}.

\bibitem[Gower(1971)]{Gower1971}
Gower, J.
\newblock {A general coefficient of similarity and some of its properties}.
\newblock \emph{Biometrics}, 27\penalty0 (4):\penalty0 857--871, 1971.

\bibitem[Grotzinger \& Witzgall(1984)Grotzinger and Witzgall]{Grotzinger1984}
Grotzinger, S. and Witzgall, C.
\newblock {Projections onto order simplexes}.
\newblock \emph{Applied Mathematics and Optimization}, 270\penalty0
  (12):\penalty0 247--270, 1984.

\bibitem[Guidotti et~al.(2018)Guidotti, Monreale, Ruggieri, Turini, Giannotti,
  and Pedreschi]{Guidotti2018}
Guidotti, R., Monreale, A., Ruggieri, S., Turini, F., Giannotti, F., and
  Pedreschi, D.
\newblock {A survey of methods for explaining black box models}.
\newblock \emph{ACM Computing Surveys}, 51\penalty0 (5):\penalty0 93:1--93:42,
  2018.

\bibitem[Jeroslow \& Lowe(1984)Jeroslow and Lowe]{Jeroslow1984}
Jeroslow, R. and Lowe, J.
\newblock {Modelling with integer variables}.
\newblock \emph{Mathematical Programming Study}, 22:\penalty0 167--184, 1984.

\bibitem[Kanamori et~al.(2020)Kanamori, Takagi, Kobayashi, and
  Arimura]{Kanamori2020}
Kanamori, K., Takagi, T., Kobayashi, K., and Arimura, H.
\newblock {DACE: Distribution-aware counterfactual explanation by mixed-integer
  linear optimization}.
\newblock \emph{Proceedings of the Twenty-Ninth International Joint Conference
  on Artificial Intelligence, IJCAI-20}, pp.\  2855--2862, 2020.

\bibitem[Karimi et~al.(2020{\natexlab{a}})Karimi, Barthe, Balle, and
  Valera]{Karimi2020a}
Karimi, A.-H., Barthe, G., Balle, B., and Valera, I.
\newblock {Model-agnostic counterfactual explanations for consequential
  decisions}.
\newblock In Chiappa, S. and Calandra, R. (eds.), \emph{Proceedings of the
  Twenty Third International Conference on Artificial Intelligence and
  Statistics}, volume 108 of \emph{Proceedings of Machine Learning Research},
  pp.\  895--905. PMLR, 2020{\natexlab{a}}.

\bibitem[Karimi et~al.(2020{\natexlab{b}})Karimi, Barthe, Sch{\"{o}}lkopf, and
  Valera]{Karimi2020}
Karimi, A.-H., Barthe, G., Sch{\"{o}}lkopf, B., and Valera, I.
\newblock {A survey of algorithmic recourse: definitions, formulations,
  solutions, and prospects}.
\newblock \emph{arXiv:2010.04050}, 2020{\natexlab{b}}.

\bibitem[Karimi et~al.(2020{\natexlab{c}})Karimi, Sch{\"{o}}lkopf, and
  Valera]{Karimi2020b}
Karimi, A.~H., Sch{\"{o}}lkopf, B., and Valera, I.
\newblock {Algorithmic recourse: From counterfactual explanations to
  interventions}.
\newblock \emph{arXiv:2002.06278}, 2020{\natexlab{c}}.

\bibitem[Liu et~al.(2008)Liu, Ting, and Zhou]{Liu2008a}
Liu, F., Ting, K., and Zhou, Z.-H.
\newblock {Isolation forest}.
\newblock In \emph{2008 Eighth IEEE International Conference on Data Mining},
  pp.\  413--422, 2008.

\bibitem[Lucic et~al.(2019)Lucic, Oosterhuis, Haned, and de~Rijke]{Lucic2019}
Lucic, A., Oosterhuis, H., Haned, H., and de~Rijke, M.
\newblock {FOCUS: Flexible optimizable counterfactual explanations for tree
  ensembles}.
\newblock \emph{arXiv:1911.12199}, 2019.

\bibitem[Mahajan et~al.(2019)Mahajan, Tan, and Sharma]{Mahajan2019}
Mahajan, D., Tan, C., and Sharma, A.
\newblock Preserving causal constraints in counterfactual explanations for
  machine learning classifiers.
\newblock \emph{arXiv:1912.03277}, 2019.

\bibitem[Mothilal et~al.(2020)Mothilal, Sharma, and Tan]{Mothilal2020}
Mothilal, R., Sharma, A., and Tan, C.
\newblock {Explaining machine learning classifiers through diverse
  counterfactual explanations}.
\newblock In \emph{Proceedings of the 2020 Conference on Fairness,
  Accountability, and Transparency}, FAT*'20, pp.\  607--617, New York, NY,
  USA, 2020. Association for Computing Machinery.

\bibitem[Rudin(2019)]{Rudin2019}
Rudin, C.
\newblock {Stop explaining black box machine learning models for high stakes
  decisions and use interpretable models instead}.
\newblock \emph{Nature Machine Intelligence}, 1\penalty0 (5):\penalty0
  206--215, 2019.

\bibitem[Russell(2019)]{Russell2019}
Russell, C.
\newblock {Efficient search for diverse coherent explanations}.
\newblock \emph{Proceedings of the Conference on Fairness, Accountability, and
  Transparency}, pp.\  20--28, 2019.

\bibitem[Tolomei et~al.(2017)Tolomei, Silvestri, Haines, and
  Lalmas]{Tolomei2017a}
Tolomei, G., Silvestri, F., Haines, A., and Lalmas, M.
\newblock {Interpretable predictions of tree-based ensembles via actionable
  feature tweaking}.
\newblock In \emph{Proceedings of the 23rd ACM SIGKDD International Conference
  on Knowledge Discovery and Data Mining}, pp.\  465--474, New York, NY, 2017.

\bibitem[Ustun et~al.(2019)Ustun, Spangher, and Liu]{Ustun2019}
Ustun, B., Spangher, A., and Liu, Y.
\newblock {Actionable recourse in linear classification}.
\newblock In \emph{Proceedings of the Conference on Fairness, Accountability,
  and Transparency}, pp.\  10--19, 2019.

\bibitem[Venkatasubramanian \& Alfano(2020)Venkatasubramanian and
  Alfano]{Venkatasubramanian2020}
Venkatasubramanian, S. and Alfano, M.
\newblock {The philosophical basis of algorithmic recourse}.
\newblock \emph{Proceedings of the 2020 Conference on Fairness, Accountability,
  and Transparency}, pp.\  284--293, 2020.

\bibitem[Verma et~al.(2020)Verma, Dickerson, and Hines]{Verma2020}
Verma, S., Dickerson, J., and Hines, K.
\newblock {Counterfactual explanations for machine learning: A review}.
\newblock \emph{arXiv:2010.10596}, 2020.

\bibitem[Vidal \& Schiffer(2020)Vidal and Schiffer]{Vidal2020a}
Vidal, T. and Schiffer, M.
\newblock {Born-again tree ensembles}.
\newblock In III, H.~D. and Singh, A. (eds.), \emph{Proceedings of the 37th
  International Conference on Machine Learning}, volume 119, pp.\  9743--9753,
  Virtual, 2020. PMLR.

\bibitem[Wachter et~al.(2018)Wachter, Mittelstadt, and Russell]{Wachter2018}
Wachter, S., Mittelstadt, B., and Russell, C.
\newblock {Counterfactual explanations without opening the black box: Automated
  decisions and the GDPR}.
\newblock \emph{Harvard Journal of Law {\&} Technology}, 31:\penalty0 841,
  2018.

\bibitem[Wolsey(2020)]{Wolsey2020}
Wolsey, L.~A.
\newblock \emph{{Integer Programming}}.
\newblock John Wiley {\&} Sons, Hoboken, NJ, 2020.
\newblock ISBN 9781119606475.

\end{thebibliography}
\end{document}